\documentclass[journal, onecolumn]{IEEEtran}

\usepackage{amsmath}
\usepackage{url}
\usepackage{amssymb}

\usepackage{lineno}

\usepackage{mathtools}
\usepackage{bm}
\usepackage{subcaption}
\usepackage{algorithm}
\usepackage{algpseudocode}
\usepackage{multirow}
\usepackage{hhline}

\usepackage{color}
\usepackage{xcolor,colortbl}
\usepackage[normalem]{ulem}

\usepackage{graphicx}
\usepackage{tabularx}

\definecolor{Gray}{gray}{0.85}

\algnewcommand{\IfThenElse}[3]{
  \State \algorithmicif\ #1\ \algorithmicthen\ #2\ \algorithmicelse\ #3}

\begin{document}

\title{Use of BIM Data as Input and Output for Improved Detection of Lighting Elements in Buildings}
\author{Francisco Troncoso-Pastoriza, Pablo Egu\'{i}a-Oller, Rebeca P. Díaz-Redondo, Enrique Granada-\'{A}lvarez
\thanks{Francisco Troncoso-Pastoriza, Pablo Egu\'{i}a-Oller and Enrique Granada-\'{A}lvarez are with School of Industrial Engineering, University of Vigo, Campus Universitario, 36310 Vigo (Spain)}
\thanks{Rebeca P. Díaz-Redondo (rebeca@det.uvigo.es) is with School of Telecommunication Engineering, University of Vigo, Campus Universitario, 36310 Vigo (Spain)}
}

\maketitle

\begin{abstract}
This paper introduces a complete method for the automatic detection, identification and localization of lighting elements in buildings, leveraging the available building information modeling (BIM) data of a building and feeding the BIM model with the new collected information, which is key for energy-saving strategies. The detection system is heavily improved from our previous work, with the following two main contributions: (i) a new refinement algorithm to provide a better detection rate and identification performance with comparable computational resources and (ii) a new plane estimation, filtering and projection step to leverage the BIM information earlier for lamps that are both hanging and embedded. The two modifications are thoroughly tested in five different case studies, yielding better results in terms of detection, identification and localization.
\end{abstract}


\section{Introduction}
\label{intro}

Building information modeling (BIM) is ``a set of interacting policies,
processes and technologies producing a methodology to manage essential building design and project data in digital format throughout the building's lifecycle''~~\cite{Succar}. This methodology, which is increasingly investigated in the architecture, engineering and construction (AEC) industry~\cite{Sanhudo, Lu, Rahmani}, facilitates the distribution of information about all the elements of the infrastructure of a building throughout its entire lifecycle, representing its digital model as a central database~\cite{Rahmani}.

This methodology can be used in conjunction with automatic detection methods~\cite{Troncoso1,Troncoso2} both as input, to acquire important data that can be used in this process, and as output, providing the relevant information of the actual state and conditions of the building. Knowing these real conditions of the building is an important factor for reducing energy consumption, which accounts for approximately 40\% of total energy consumption worldwide, with a growing trend that is not expected to decrease in the short term~\cite{Lombard}.

Lighting is one of the most important factors in this consumption, representing approximately 19\% of the total electricity used in the world~\cite{iea2}, with approximately one-third of the electricity in buildings being used for artificial lighting~\cite{Soori, Lombard, Baloch}. In fact, lighting is one of the main issues in the analysis of multiple performance criteria in BIM~\cite{Lu, Rahmani, Welle}. Therefore, knowing the real state of the lighting elements in a building is critical to perform energy conservation measures (ECMs)~\cite{Baloch} to reduce energy use and costs~\cite{Soori}. However, one of the main problems is the absence of accurate information~\cite{Lu}. New methodologies have been proposed to solve this problem using automatic detection techniques based on computer vision~\cite{Vilariño,Troncoso1,Troncoso2}.

Computer vision is a technology that is widely used to automate the object recognition process and has already successfully been used in the lighting industry: Elvidge et al.~\cite{Elvidge} analyzed the optimal spectral bands to identify lighting types, obtaining four major indices to measure lighting efficiency; Liu et al.~\cite{Liu2} presented an imaging sensor-based LED lighting system, yielding a more precise lighting control; and Ng et al. [15] proposed a lighting inspection system based on a practical and fast approach using computer vision and imaging processing tools.

The wide variety of methods for object detection can be classified into two categories: image-based~\cite{Viksten} and model-based~\cite{Tombari}. Different methods for these two categories have been already comprehensively presented in a previous article~\cite{Troncoso1}, with the model-based category being the most adequate for the detection of lighting elements due to the untextured nature of this kind of object~\cite{Troncoso1}. Among these methods, chamfer matching~\cite{Barrow,Borgefors} has been used for shape-based object detection, leveraging the information of the edges in the image, one of the most important low-level image features~\cite{Borgefors}. Several variations and improvements of this method are presented in~\cite{Troncoso1}, including oriented chamfer matching (OCM)~\cite{Shotton}, which includes an additional channel specifically for the orientation information, and fast directional chamfer matching (FDCM)~\cite{fdcm}, which performs a fast matching in a joint location/orientation space. Based on the FDCM, direct directional chamfer optimization (D$^2$CO) was proposed~\cite{D2CO} for object registration, refining the position and orientation of the object using a nonlinear optimization process. This object registration method has been already used inside a framework to detect lighting elements in buildings~\cite{Troncoso1,Troncoso2}.

\begin{figure}[ht]
\centering
\includegraphics[width=0.95\linewidth]{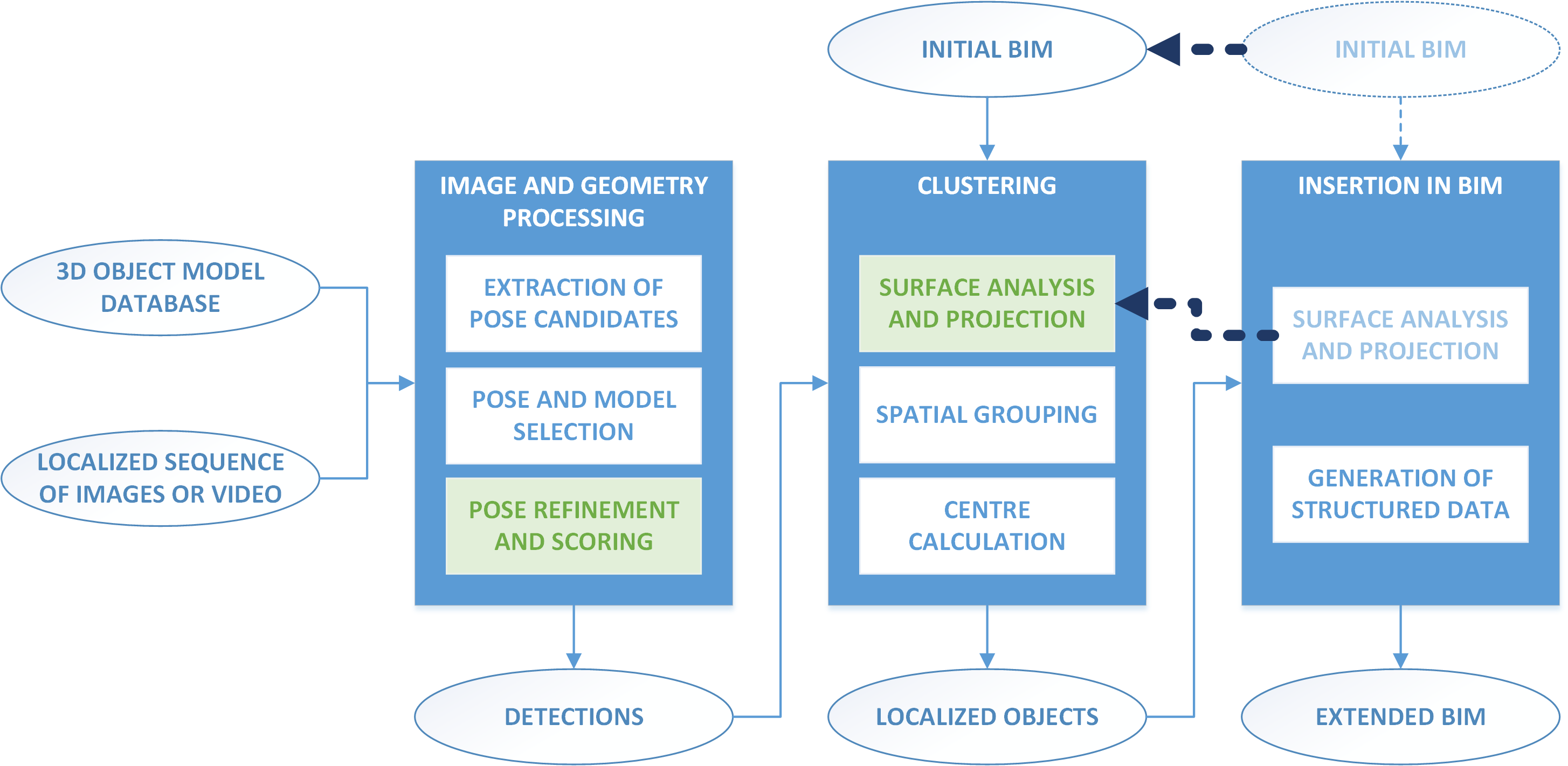}
\caption{Updated BIM generation process.}
\label{fig:general}
\end{figure}

In this work, we propose two modifications to the workflow presented in our previous work~\cite{Troncoso1,Troncoso2}: (i) a new refinement algorithm based on D$^2$CO to improve the performance of the detection and (ii) a new surface projection method that leverages the BIM information earlier, producing more accurate results and allowing for the use of BIM data for both embedded and hanging lamps. The modified scheme is presented in Figure~\ref{fig:general}, with the two improvements indicated with blue arrows and green boxes.

The new methodology is explained in two different sections: Section~\ref{sec:method-opt} for the refinement algorithm and Section~\ref{sec:method-cls} for the surface projection method. The description of the experimental system used for the validation is described in Section~\ref{sec:exp}, and the results are presented in Figure~\ref{sec:results}. Finally, the conclusions obtained from these results are included in Section~\ref{sec:conclusions}.

\section{Improved refinement step}
\label{sec:method-opt}

\begin{figure}[ht]
\centering
\includegraphics[width=0.95\linewidth]{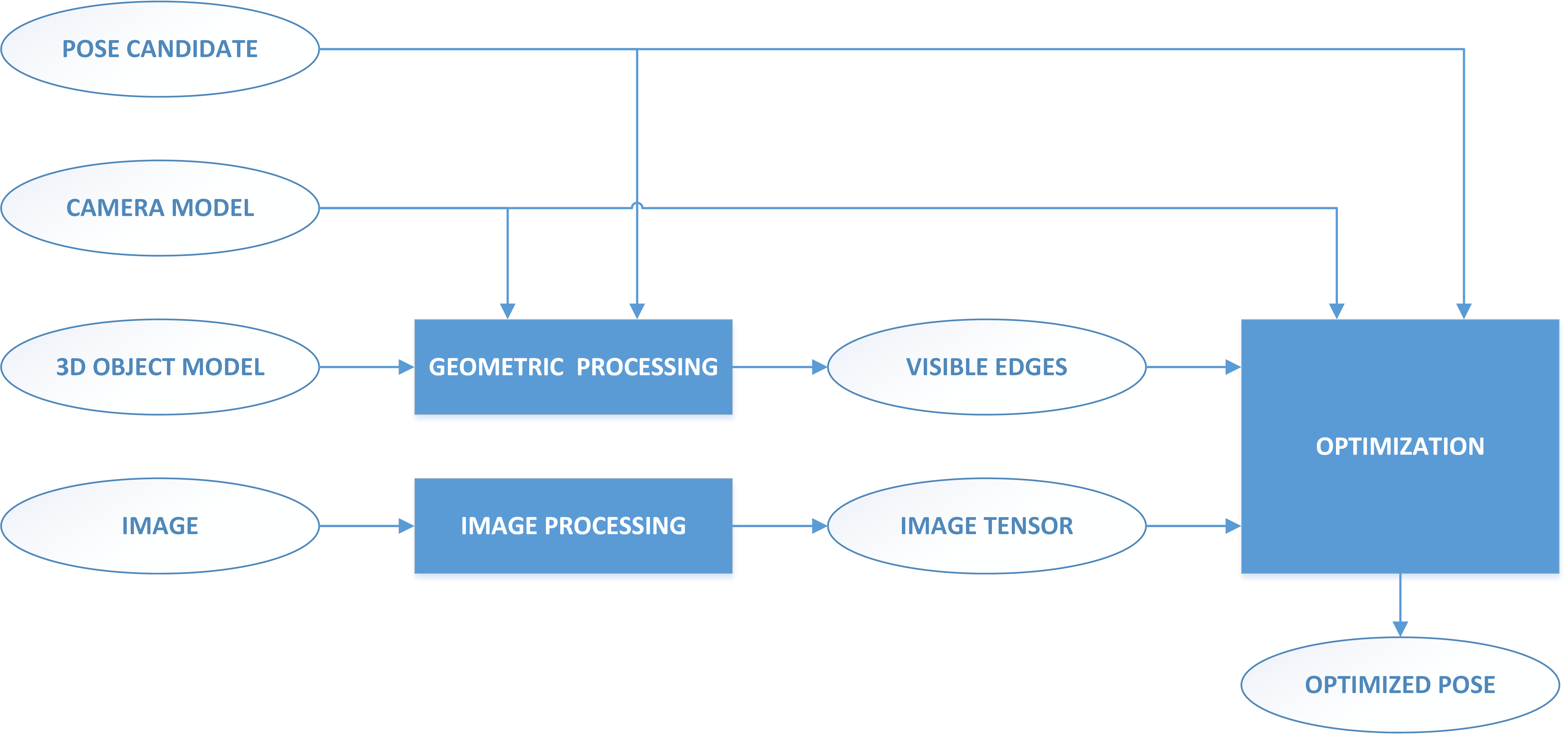}
\caption{General overview of the complete detection process.}
\label{fig:general-opt}
\end{figure}

In this section, we describe our proposed method to improve the detection and identification performance of the system, \emph{direct directional chamfer optimization with integral tensor} (D$^2$CO-IT), based on the D$^2$CO~\cite{D2CO} that we used in previous works~\cite{Troncoso1,Troncoso2}. In addition to the main optimization, we also include the previous steps that are needed to provide the necessary input and that comprise two main parts: the analysis of 3D geometric information based on the current camera configuration and 3D mesh of the candidate object and the analysis of the image information. The entire procedure is depicted in Figure~\ref{fig:general-opt}.

\subsection{Geometric processing: Extraction of visible edges}

\begin{figure}[ht]
\centering
\includegraphics[width=\linewidth]{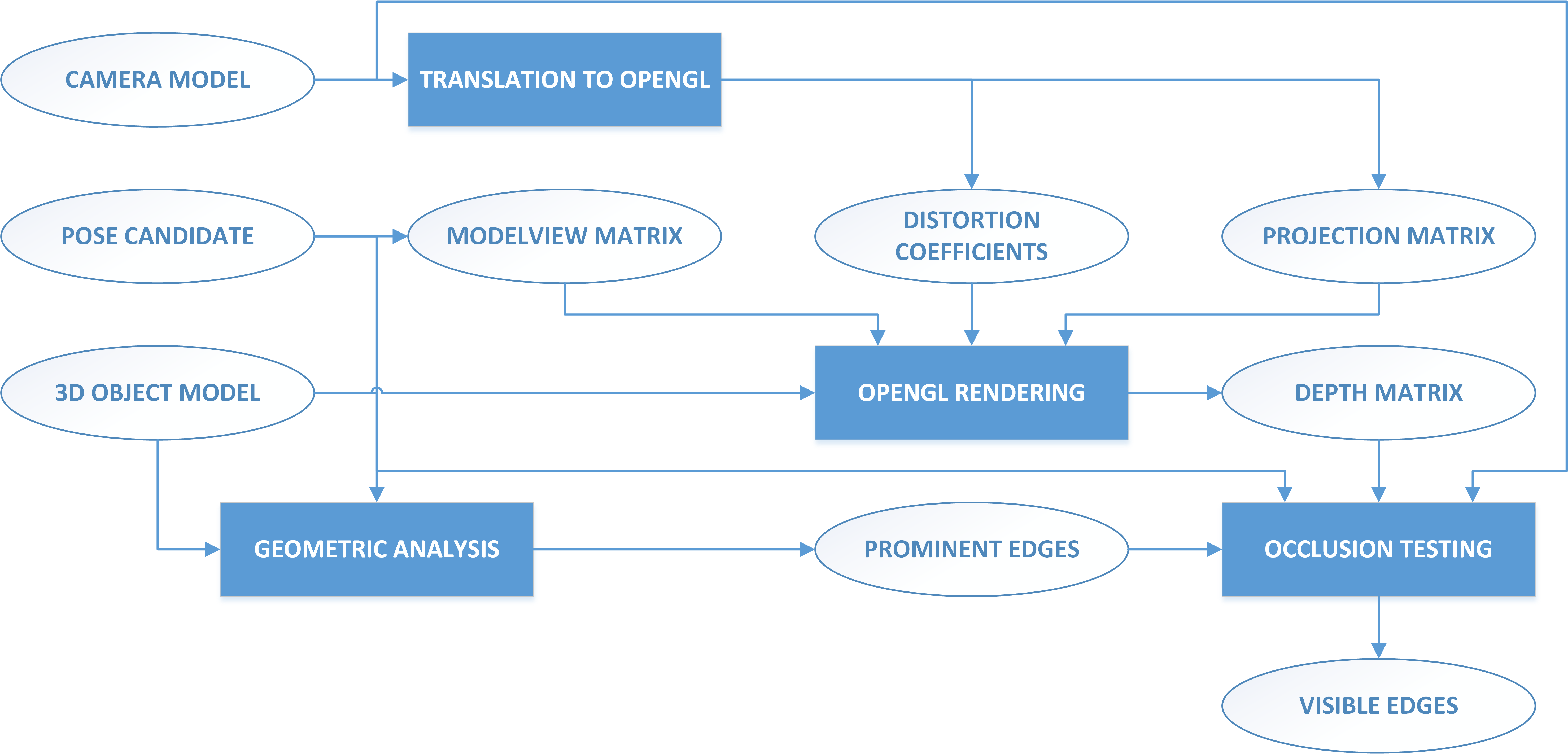}
\caption{Diagram of the operations performed in the geometric processing step.}
\label{fig:3d}
\end{figure}

Figure~\ref{fig:3d} shows the main operations that are performed in this step: geometric analysis of edges, extraction of depth information and occlusion testing. This procedure has barely changed from the work presented in~\cite{Troncoso1}, but we include a brief description of all the steps for the sake of completeness.

We use the term \emph{pose} to denote a rigid transformation of an object, i.e., a direct Euclidean isometry defined as a $4 \times 4$ matrix in $\textbf{SE(3)}$ with the structure presented in Eq.~\eqref{eq:modelview}:

\begin{align}\label{eq:modelview}
\bm M =& \left(\begin{array}{c|c}
\bm R & \bm t \\
\hline
\bm 0_{1x3} & 1
\end{array}\right),
\end{align}
where $\bm R \in \textbf{SO(3)}$ is a $3 \times 3$ skew-symmetric rotation matrix, and $\bm t \in \mathbb{R}^3$ is a translation vector. We can also represent a \emph{pose} as an element in the corresponding Lie algebra $\mathfrak{se}\textbf{(3)}$, i.e., a 6-dimensional vector comprising a vector $\bm t \in \mathbb{R}^3$ that determines the translation and a vector $\bm \omega \in \mathfrak{so}\textbf{(3)}$ that determines the orientation.

Changing between both representations can be done by means of the corresponding exponential and logarithmic maps, which are given by the Rodrigues rotation formula for $\textbf{SO(3)}$. We use the Lie group matrix representation for the intermediate operations and the Lie algebra vector representation as the input variables in the optimization problem.


The first step comprises a geometric analysis to obtain the prominent edges of a 3D mesh from a given camera pose and model. This includes all edges that meet one of the following two conditions: it is sharp or is part of the outline.
This process is very similar to the one presented in~\cite{Troncoso1}, with the exception that in this case we use only the interior angle for the sharpness test since, in our experiments, the sharp edges going inside the polyhedron rarely correspond to sharp differences in the image. This modified procedure is described below.


Let $\mathcal{E}$ be an edge in a polyhedron going from points $\bm{p_a}$ to $\bm{p_b}$. Let $\mathcal{F}_a$ and $\mathcal{F}_b$ be the two faces adjacent to $\mathcal{E}$ with normals $\bm{n_a}$ and $\bm{n_b}$, respectively. Let $\bm{v_b}$ be a vector pointing from any point in $\mathcal{E}$ to any other point in $\mathcal{F}_b$ not in $\mathcal{E}$. Then, $\mathcal{E}$ is considered sharp for a predefined threshold $\mathcal{S}$ if both Eqs.~\eqref{eq:sharp1} and \eqref{eq:sharp2} hold:

\begin{align}
\label{eq:sharp1} \bm{n_a} \cdot \bm{n_b} < \mathcal{S}, \\
\label{eq:sharp2} \bm{n_a} \cdot \bm{v_b} < 0,
\end{align}
i.e., the interior angle between its two adjacent faces is lower than a certain value. 
In our experiments, we use a value for $\mathcal{S}$ corresponding to 140 degrees of angular difference between adjacent faces since this value captures all the relevant edges in the image.


To obtain depth information to later use in the occlusion tests, we render the object mesh in OpenGL~\cite{OpenGL} using the same parameters as the candidate pose, and we later read the values of the depth buffer. We need to perform some transformations on the input data to properly configure the OpenGL scene as mentioned in~\cite{Troncoso1}.

Once we have obtained the prominent edges from the geometric analysis and the depth information from the OpenGL depth buffer, we use all that information to extract the final visible edges. This method is based on a discretization of the input edges, testing the occlusion for every point of every edge and compacting the resulting visible points into edges.


\subsection{Image processing: integral distance transform tensor}

For the optimization process, we need to obtain the integral distance transform tensor (IDT3$_V$) as described in the work of Liu \emph{et al.} \cite{fdcm}. The main difference between our proposed optimization and the reference method D$^2$CO~\cite{D2CO} is that we leverage the IDT3$_V$ instead of the distance transform tensor (DT3$_V$) to improve the performance of the refinement process while keeping comparable computational needs.

\paragraph{Edge extraction and distance transform}

The first step consists of obtaining the edges from the image; for this, we use the line segment detector (LSD)~\cite{lsd}, with the following internal parameters: $\rho = 1.83$, $\tau = 22.5$ and $\epsilon = 1$. The only difference with respect to the default values presented in~\cite{lsd} is $\rho$, which is adjusted to obtain more edges, even when the error might be a bit higher; this is especially important to obtain enough data in images with difficult lighting conditions. This information is used to generate the DT3$_V$ as described in~\cite{Liu}. Hereafter, we include a brief description of this step for the sake of completeness: first, the orientation space is quantized in $N_{\theta} = 60$ bins as in the original work~\cite{Liu} to reduce quantization artifacts\footnote{In our experiments, values higher than 60 for $N_{\theta}$ result in negligible improvements while imposing a higher computational cost.}, and each of the edges is included in the corresponding binary image associated with its quantized orientation $Q(\theta)$; then, a distance transform is applied on each of the binary images; and finally, the distances between different orientations are included using a forward and a backward recursion with a penalty factor of $\lambda_{\theta} = 100$. Then, a smoothing operation is performed on the DT3$_V$, consisting of a Gaussian filter along the orientation dimension, as suggested in~\cite{D2CO}.

\begin{figure}[ht]
\centering
\includegraphics[width=0.5\linewidth]{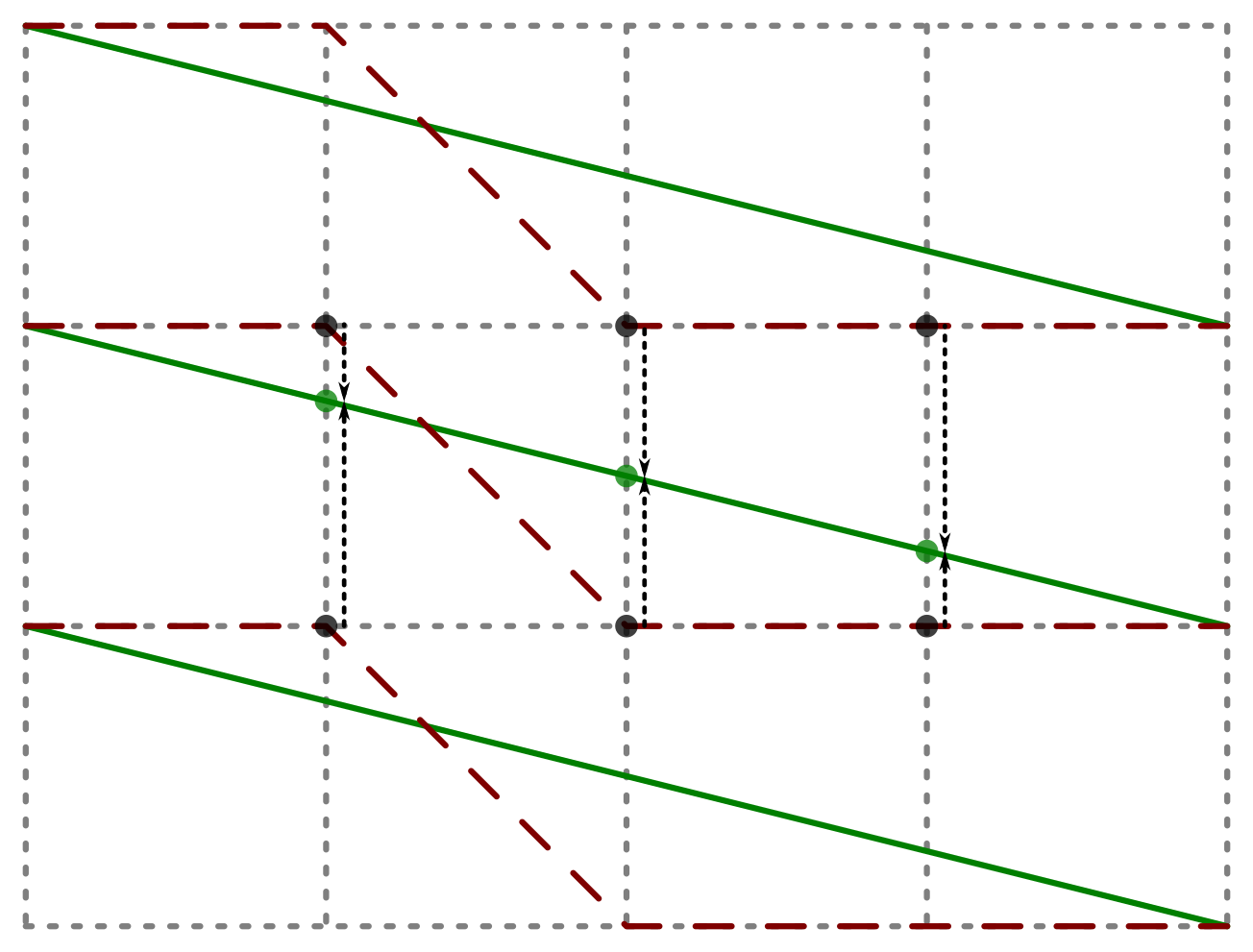}
\caption{Integration lines on the image. Green solid line: real integration line, matching the one with interpolated values; red dashed line: real integration with the direct method.}
\label{fig:integral_interp}
\end{figure}

The final IDT3$_V$ is obtained by integrating each of the orientation images in the DT3$_V$ along the corresponding direction. In this step, we perform a linear interpolation between adjacent pixel values, as the theoretical integration line does not match exactly with the pixel positions, as shown in Figure~\ref{fig:integral_interp}.

\subsubsection{Error analysis of the integration}
\label{subsec:error}

\begin{figure}[ht]
\centering
\includegraphics[width=0.5\linewidth]{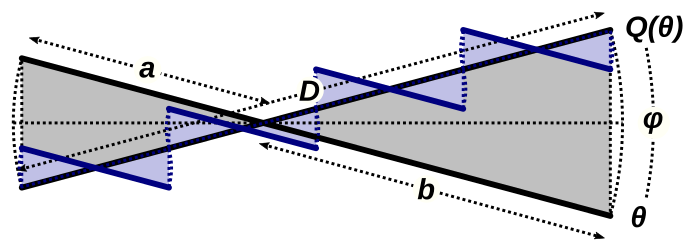}
\caption{Diagram of an edge with its rotated version by an angle $\varphi$ to fit the nearest quantized orientation. The shaded area corresponds to the error associated with this operation. Dividing the edge reduces the total error of the combined fragments, as shown in blue.}
\label{fig:integral_edge_error}
\end{figure}

Performing the integration step highly increases the evaluation speed of edge distances, but it also comes with an intrinsic error in the results with respect to a direct evaluation. This is due to the fact that to evaluate a distance on the IDT3$_V$, the orientation of the edge has to be fixed to a quantized value, fitting in an integration line in the image. This implies that the edge has to be rotated by an angle $\varphi$, as illustrated in Figure \ref{fig:integral_edge_error}, which means that the individual points are not in the exact position of the original edge. We can obtain an upper bound for this error based on the idea that the images in the DT3$_V$ are distance-transformed.

Let $\bm{p_a}$ and $\bm{p_b}$ be two pixels in a distance-transformed image; let $v(\bm{p_a})$ and $v(\bm{p_b})$ be their values on the image and $d(\bm{p_a}, \bm{p_b})$ the distance between them. Then, Eq.~\eqref{eq:pixel_bound} is true:

\begin{align}\label{eq:pixel_bound}
|v(\bm{p_a}) - v(\bm{p_b})| \leq d(\bm{p_a}, \bm{p_b}) .
\end{align}

Let $e_\theta$ be an edge of length $D$ between points $\bm{e_\theta^{(a)}}$ and $\bm{e_\theta^{(b)}}$. Let $e_{Q(\theta)}$ be a rotated version of $e_\theta$, by an angle $\varphi$, with a rotation center that is in $e_\theta$ at a distance $a$ from $\bm{e_\theta^{(a)}}$ and a distance $b = D - a$ from $\bm{e_\theta^{(b)}}$. Then, based on Eq.~\eqref{eq:pixel_bound}, an upper bound for the quantization error is defined in Eq.~\eqref{eq:error_bound}:

\begin{align}\label{eq:error_bound}
\nonumber
\varepsilon \leq& \int_0^a 2x \sin (\varphi / 2) \,\text{d}x + \int_0^b 2x \sin (\varphi / 2) \,\text{d}x \\
\nonumber
=& (a^2 + b^2) \sin (\varphi / 2) \\
=& (2a^2 - 2Da + D^2) \sin(\varphi / 2).
\end{align}

Therefore, the value of $a$ that minimizes the error is $a = D / 2$.
This result means that the center of rotation has to be in the middle of the edge to obtain the minimum error.

Using this value for $a$ and given that the maximum value of the rotation is $\varphi_{max} = \Delta_\theta / 2$, with $\Delta_\theta$ being the angle between consecutive quantized orientations, we obtain the final value of an upper bound for the error in Eq.~\eqref{eq:error}:

\begin{align}\label{eq:error}
\varepsilon_{max} = \frac{D^2}{4} \sin\left(\frac{\Delta_\theta}{4}\right) .
\end{align}

This bound is, thus, proportional to $D^2$. If we divide the desired edge into $N$ different equally sized segments, the boundary decreases as shown in Eq.~\eqref{eq:error_n}:

\begin{align}\label{eq:error_n}
\varepsilon_{max}^{(N)} = \sum_{i=1}^N \frac{(D/N)^2}{4} \sin\left(\frac{\Delta_\theta}{4}\right) = \frac{\varepsilon_{max}}{N} .
\end{align}

Therefore, to reduce the potentially higher values of this quantization error in large edges, in our experiments, we perform a discretization of the edges of the model based on a step value that is proportional to the largest edge in the model.

\subsection{Optimization}

The main goal of the method is to improve the initial candidate pose based on the edge information in the image. For this, we use a nonlinear optimization process that minimizes the directional chamfer distance based on the values of the IDT3$_V$.

Let $\bf\Pi$ be a pose for an object, $\kappa$ the set of parameters for a given camera model, and $\bm{\mathcal{E}} = \{\mathcal{E}_i\}$, with $i = 1, \ldots, N$, the set of edges for the object model. Let $\rho(\mathcal{E}_i, \bf\Pi, \kappa)$ be a function that returns the projected 2D edge from given 3D edge, pose and camera parameters, and $\mathcal{D}_\mathfrak{T}(e)$ a function that returns the distance for a given 2D edge using the values of the integral tensor $\mathfrak{T}$. Then, the optimization problem tries to minimize the error presented in Eq.~\eqref{eq:le_error}:

\begin{align}\label{eq:le_error}
E({\bf\Pi}) = \frac{1}{2} \sum_{i=1}^{N} \mathcal{D}_{\text{IDT3$_V$}}(\rho(\mathcal{E}_i, \bf\Pi, \kappa))^2 .
\end{align}

We compute the derivatives of the error function, $\nabla E$, to perform the minimization. In the case of direct accesses to the integral tensor values, we compute the numerical derivatives, as the tensor is only defined at discrete points in $x$, $y$ and $\theta$.
In the following section, we define the distance function $\mathcal{D}_\mathfrak{T}(e)$.

\subsubsection{Distance calculation}

Using the IDT3$_V$, we can only retrieve values at discrete points; therefore, to obtain a continuous function, we perform a series of interpolations. Let $f$ be a function defined at discrete points in $\mathbb{Z}$ for a given variable $x$. Then, we define the interpolation operation $\Upsilon$ on the values of $f$ for the variable $x$ as in Eq.~\eqref{eq:interp}:

\begin{align}\label{eq:interp}
\Upsilon_x(f) = (1 - (x - \lfloor x \rfloor)) f(\lfloor x \rfloor) + (x - \lfloor x \rfloor) f(\lfloor x \rfloor + 1) ,
\end{align}
with $\lfloor \cdot \rfloor$ being the floor function.

\begin{figure}[ht]
\centering
\includegraphics[width=\linewidth]{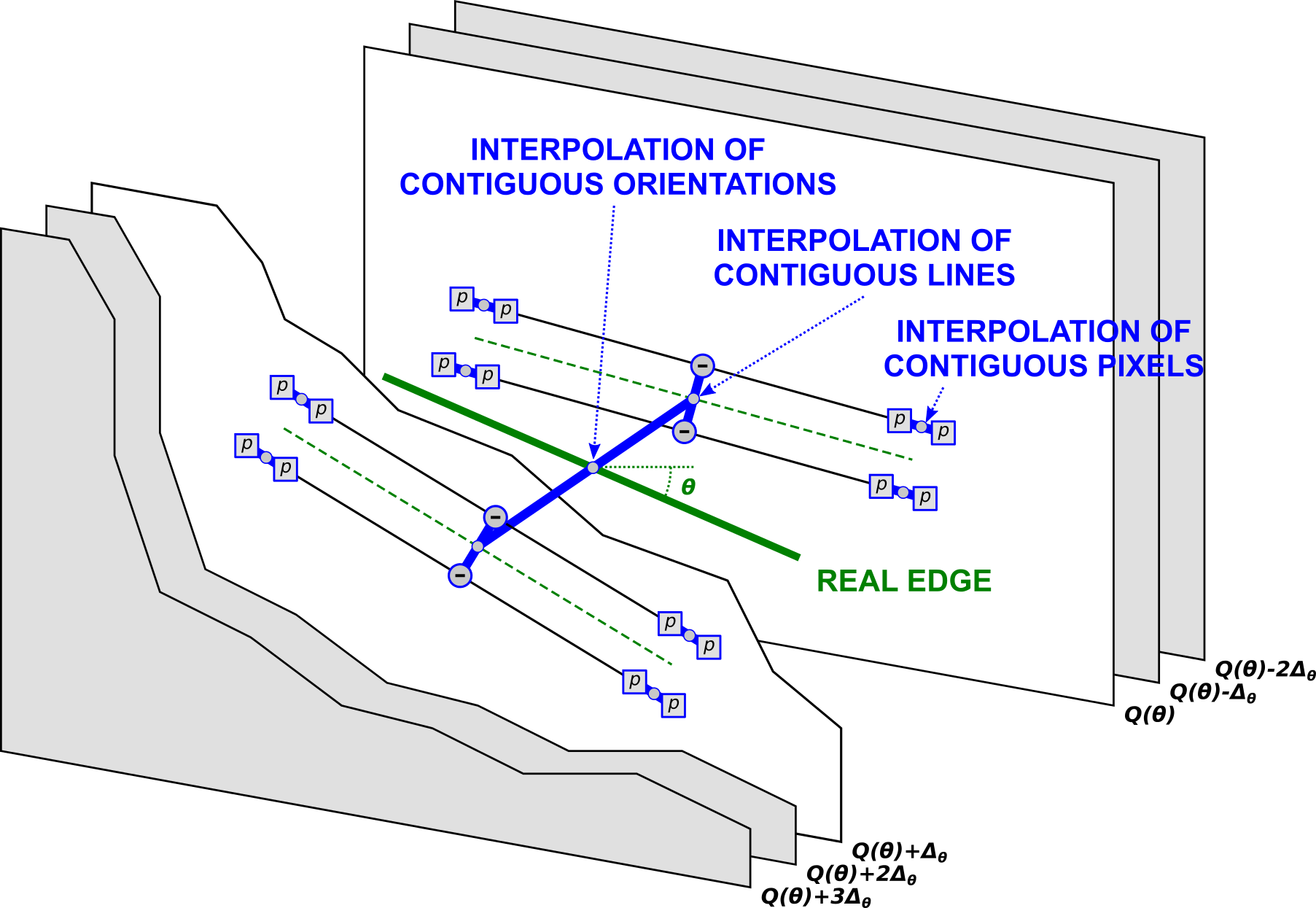}
\caption{Diagram of the interpolations performed to retrieve the final edge distance value. The squares with $p$ represent pixel values, the blank dots indicate interpolated values, and the circles with a ``$-$'' sign represent a value resulting from a subtraction operation.}
\label{fig:interpolations}
\end{figure}

As shown in Section~\ref{subsec:error}, the best point of connection between different orientations is the center of the projected edge; thus, for a given edge $e$ comprised of points $\bm{e^{(a)}}$ and $\bm{e^{(b)}}$, we first perform a transformation to define the edge in terms of $\{\bm{e^{(c)}}, r, \theta\}$, i.e., their center, half-length and orientation. From this alternate representation, we can obtain the corresponding integration line $\mathfrak{l}(e_x^{(c)}, e_y^{(c)}, \theta)$ and pixel in that line $\mathfrak{p}(e_x^{(c)}, e_y^{(c)}, \theta)$ of the edge center for a given orientation $\theta$.
Finally, the distance of an edge $e = \{\bm{e^{(c)}}, r, \theta\}$ on the tensor $\mathfrak{T}$ is defined in Eqs.~\eqref{eq:tensor_dist}, \eqref{eq:tensor_dist_b} and \eqref{eq:tensor_dist_c}:

\begin{align}
\label{eq:tensor_dist} \mathcal{D}_\mathfrak{T}(e_x^{(c)}, e_y^{(c)}, r, \theta) =& \Upsilon_\theta(\Delta_\mathfrak{T}(\mathfrak{l}(e_x^{(c)}, e_y^{(c)}, \theta), \mathfrak{p}(e_x^{(c)}, e_y^{(c)}, \theta), r, \theta)) ,\\
\label{eq:tensor_dist_b} \Delta_\mathfrak{T}(l, p, r, \theta) =& \Upsilon_l(\delta_\mathfrak{T}(l, p, r, \theta)) ,\\
\label{eq:tensor_dist_c} \delta_\mathfrak{T}(l, p, r, \theta) =& \Upsilon_p(\mathfrak{T}(l, p+r, \theta)) - \Upsilon_p(\mathfrak{T}(l, p-r, \theta)) ,
\end{align}
with $\mathfrak{T}(l, p, \theta)$ being the value of $\mathfrak{T}$ for the pixel $p$ of the integration line $l$ in the slice corresponding to the orientation $\theta$. The complete set of operations is depicted in Figure~\ref{fig:interpolations} and enumerated in detail in Algorithms~\ref{alg:edgedist}~to~\ref{alg:precomp}.

\begin{algorithm}[!ht]
\caption{Edge distance}
\label{alg:edgedist}
\begin{algorithmic}[1]
\Require
\Statex $\mathfrak{T}$: values of an integral tensor
\Statex $\bm a$, $\bm b$: edge vertices
\Ensure
\Statex $D$: edge distance for the given tensor
\Statex
\State $\bm \Delta \gets \bm b - \bm a$
\State $d \gets \|\bm \Delta\|$
\State $\bm c \gets 0.5 \cdot (\bm a + \bm b)$
\State $z \gets \Omega(\bm \Delta) * \text{len}(\mathfrak{T}) / \pi$
\State $r \gets 0.5 \cdot d$
\State $z_a \gets \lfloor z \rfloor$
\State $z_b \gets z_a + 1$
\State $d_z \gets z - z_a$
\State $P_a \gets $\Call{Precomputed}{$z_a$} \Comment Algorithm~\ref{alg:precomp}
\State $v_a \gets$\Call{GetImageValue}{$\mathfrak{T}[z_a]$, $\bm c$, $r$, $P_a$} \Comment Algorithm~\ref{alg:getimval}
\State $P_b \gets $\Call{Precomputed}{$z_b$} \Comment Algorithm~\ref{alg:precomp}
\State $v_b \gets$\Call{GetImageValue}{$\mathfrak{T}[z_b]$, $\bm c$, $r$, $P_b$} \Comment Algorithm~\ref{alg:getimval}
\State $D \gets$ $(v_a + (v_b - v_a) \cdot d_z) / d$
\end{algorithmic}
\end{algorithm}

\begin{algorithm}[!ht]
\caption{Get image value}
\label{alg:getimval}
\begin{algorithmic}[1]
\Require
\Statex $\mathfrak{I}$: values of an integral image
\Statex $\bm c$, $r$: center and radius of edge
\Statex $P$: precomputed orientation values
\Ensure
\Statex $D$: edge distance for the given image
\Statex
\IfThenElse{$P.x_d$}{$m \gets \text{cols}(\mathfrak{I})$}{$m \gets \text{rows}(\mathfrak{I})$}
\IfThenElse{$P.x_p$}{$x \gets c_x$}{$x \gets m - 1 - c_x$}
\IfThenElse{$P.x_d$}{$p \gets x$}{$p \gets c_y$}
\IfThenElse{$P.x_d$}{$l \gets c_y - p \cdot P.d_s$}{$l \gets x - p \cdot P.d_s$}
\State $l_a \gets \lfloor l \rfloor$
\State $l_b \gets l_a + 1$
\State $d_l \gets l - l_a$
\State $p_a \gets p + d_l * P.l_p$
\State $p_b \gets p_a - P.l_p$
\State $r_p \gets r \cdot P.d_p$
\State $v_{a,1} \gets $\Call{GetPixelValue}{$\mathfrak{I}$, $l_a$, $p_a - r_p$, $P.d_s$} \Comment Algorithm~\ref{alg:getpixval}
\State $v_{a,2} \gets $\Call{GetPixelValue}{$\mathfrak{I}$, $l_a$, $p_a + r_p$, $P.d_s$} \Comment Algorithm~\ref{alg:getpixval}
\State $v_{b,1} \gets $\Call{GetPixelValue}{$\mathfrak{I}$, $l_b$, $p_b - r_p$, $P.d_s$} \Comment Algorithm~\ref{alg:getpixval}
\State $v_{b,2} \gets $\Call{GetPixelValue}{$\mathfrak{I}$, $l_b$, $p_b + r_p$, $P.d_s$} \Comment Algorithm~\ref{alg:getpixval}
\State $D \gets$ $(v_{a,2} - v_{a,1}) + ((v_{b,2} - v_{b,1}) - (v_{a,2} - v_{a,1})) \cdot d_l$
\end{algorithmic}
\end{algorithm}

\begin{algorithm}[!ht]
\caption{Get pixel value}
\label{alg:getpixval}
\begin{algorithmic}[1]
\Require
\Statex $\mathfrak{I}$: values of an integral image
\Statex $l$, $p$: pixel coordinates in line-pixel space
\Statex $d_s$: orientation ratio
\Ensure
\Statex $V$: interpolated pixel value
\Statex
\State $p_a \gets \lfloor p \rfloor$
\State $p_b \gets p_a + 1$
\State $d_p \gets p - p_a$
\State $v_a \gets \mathfrak{I}[l + \lfloor p_a \cdot d_s \rfloor, p_a]$
\State $v_b \gets \mathfrak{I}[l + \lfloor p_b \cdot d_s \rfloor, p_b]$
\State $V \gets$ $v_a + (v_b - v_a) \cdot d_p$
\end{algorithmic}
\end{algorithm}

\begin{algorithm}[!ht]
\caption{Precomputed}
\label{alg:precomp}
\begin{algorithmic}[1]
\Require
\Statex $z$: orientation index $\in [0, \text{len}(\mathfrak{T}))$
\Ensure
\Statex $P$: relevant orientation values, which can be precomputed for all the orientations outside of the optimization loop
\Statex
\State $o \gets z \cdot \pi / \text{len}(\mathfrak{T})$
\State $\bm \Delta \gets \{\cos(o), \sin(o)\}$
\State $d_a \gets |\Delta_x| $
\State $x_p \gets (\Delta_x \ge 0)$
\State $x_d \gets (d_a \ge \Delta_y)$
\IfThenElse{$x_d$}{$d_p \gets d_a$}{$d_p \gets d_y$}
\IfThenElse{$x_d$}{$d_s \gets d_y / d_a$}{$d_s \gets d_a / d_y$}
\State $l_p \gets d_a \cdot d_y$
\State $P \gets$ $\{x_p, x_d, d_p, d_s, l_p\}$
\end{algorithmic}
\end{algorithm}

\section{Improved clustering based on BIM information}
\label{sec:method-cls}

In our previous work~\cite{Troncoso2}, the BIM data of the building were used to improve the results in the last step of the process, after the clustering operation. In this work, we present a modified algorithm to leverage that valuable information before the clustering is performed to reduce the dispersion of the individual detections. Moreover, we introduce an additional step to take advantage of this method also with hanging lamps, without previous knowledge of the distance between the lamps and the ceiling.

The extraction of the BIM data can be performed as explained in~\cite{Troncoso2}, in which this extraction is exemplified with the green building XML schema~\cite{gbxml}, but the procedure can be adapted to the specific BIM format. From the coordinates of the points that comprise a building surface we can trivially obtain the plane equation that is used in the proposed algorithms.

\subsection{Plane estimation}

\begin{figure}[!ht]
\centering
\includegraphics[width=0.50\linewidth]{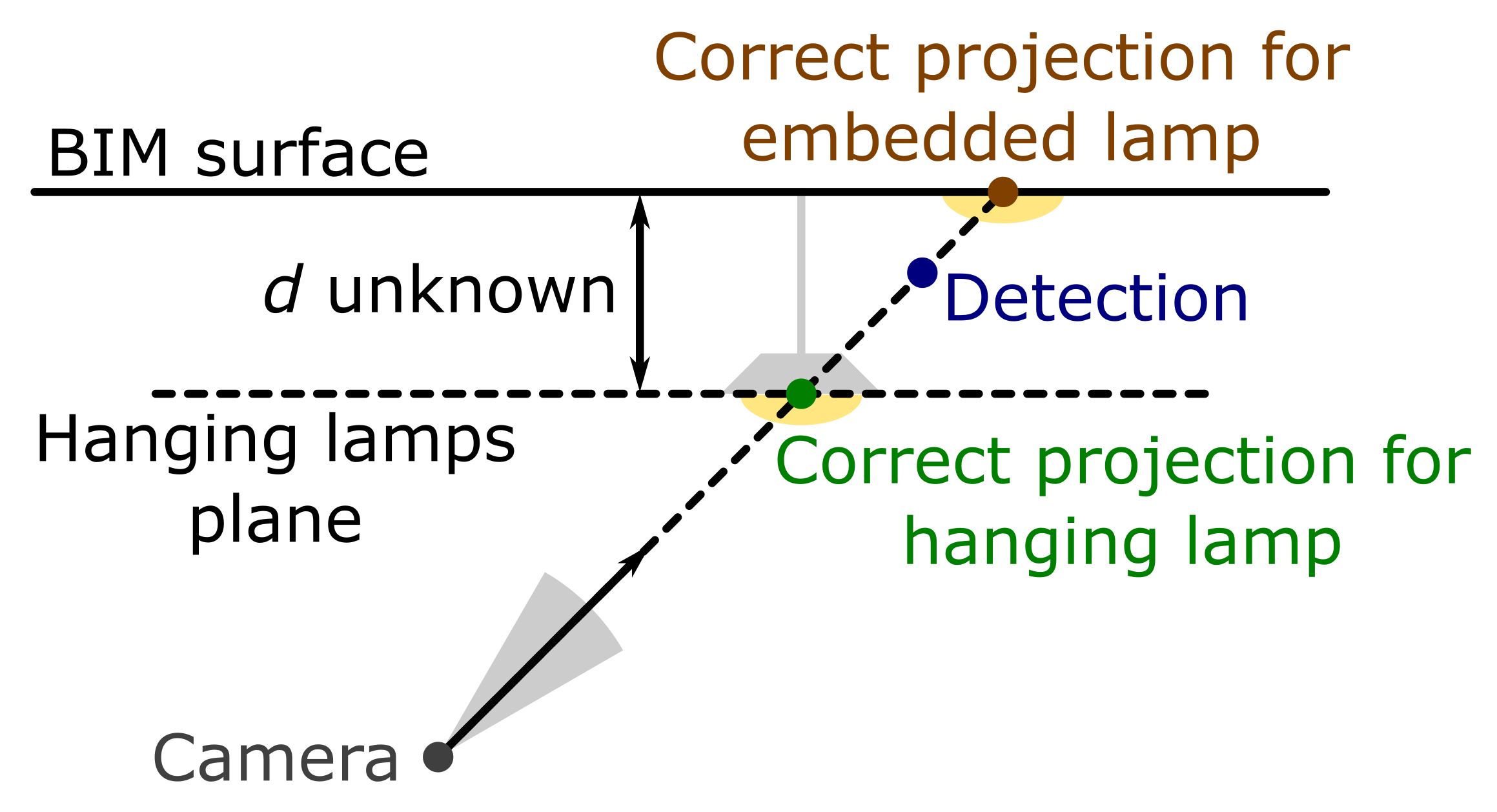}
\caption{Projection for different types of lamps.}
\label{fig:proj}
\end{figure}

Using lamp models that are embedded in the ceiling, we can correctly deduce its optimal height based on the geometric information of the building. However, this is not true if the lamps are hanging an unknown distance from the ceiling, as shown in Figure~\ref{fig:proj}. Therefore, the correct plane has to be approximated to project the detections.

We could perform this process just with the available detections, but this approach has two main drawbacks: (i) the available BIM information is not utilized, and (ii) the plane may not be correctly approximated for colinear or almost colinear points, which is typical for corridors. To overcome these problems, we propose an estimation of the plane based on the closest ceiling obtained from the geometry of the building.

Let $\mathcal{S}$ be the surface corresponding to the closest ceiling, obtained as in~\cite{Troncoso2}, for a set of $N$ detections with positions $\{p_i\}$, lying on a plane with a unit normal vector $\bm{\hat n} = (n_x, n_y, n_z)$ and an equation $\bm{\hat n} \bm{p} = d_0$. Then, the estimation consists of solving the least-squares minimization problem presented in Eq.~\eqref{eq:le_d}:

\begin{align}\label{eq:le_d}
E = \frac{1}{2} \sum_{i=1}^N \| \bm{\hat n} \bm{p_i} - d \|^2 ,
\end{align}
which provides the solution for the optimal value of $d$, shown in Eq.~\eqref{eq:dopt}:

\begin{align}\label{eq:dopt}
d_{\text{opt}} = \frac{\sum_{i=1}^N \bm{\hat n} \bm{p_i}}{N} .
\end{align}

To avoid the negative effect of outliers in the set of detections, we include this method inside the M-estimator sample consensus algorithm (MSAC)~\cite{msac}, using a sample size of 2 and a maximum distance of 30 cm to the estimated plane. The points that are farther than this threshold will be filtered for the clustering.

Finally, to project the detections, we can use the modified plane defined by Eq.~\eqref{eq:applane}:

\begin{align}\label{eq:applane}
\bm{\hat n} \bm{p} = d_{\text{opt}} .
\end{align}

\subsection{Projection of individual detections}

Individual detections can be projected to the appropriate plane before clustering. In this case, we can leverage the camera positions for this projection step.

Let $\mathcal{L}$ be a line passing through the position of a detection, $\bm{p_d}$, and the corresponding camera position, $\bm{p_c}$, at the instant in which it was captured, and let $\bm{\hat f}$ be a unit vector pointing from $\bm{p_c}$ to $\bm{p_d}$. Let $\mathcal{P}$ be the approximated plane defined in Eq.~\eqref{eq:applane}. Then, we can obtain the projected detection on the plane $\bm{p_p}$ by solving the system of linear equations in Eq.~\eqref{eq:lp_system}:

\begin{align}\label{eq:lp_system}
\begin{cases}
\mathcal{L}: \bm{p_p} = \bm{p_c} - \bm{\hat f} t \\
\mathcal{P}: \bm{\hat n} \bm{p_p} = d_{\text{opt}}
\end{cases} \rightarrow
\left(\begin{array}{c|c}
\bm I_{3x3} & \bm{\hat f}^t \\
\hline
\bm{\hat n} & 0
\end{array}\right)
\left(\begin{array}{c}\bm{p_p}\\ \hline t\end{array}\right) =
\left(\begin{array}{c}\bm{p_c}\\ \hline d_{\text{opt}}\end{array}\right) .
\end{align}

\section{Experiments}
\label{sec:exp}

\begin{figure}[!ht]
\centering
\begin{subfigure}{0.325\linewidth}
\centering
\includegraphics[width=\textwidth]{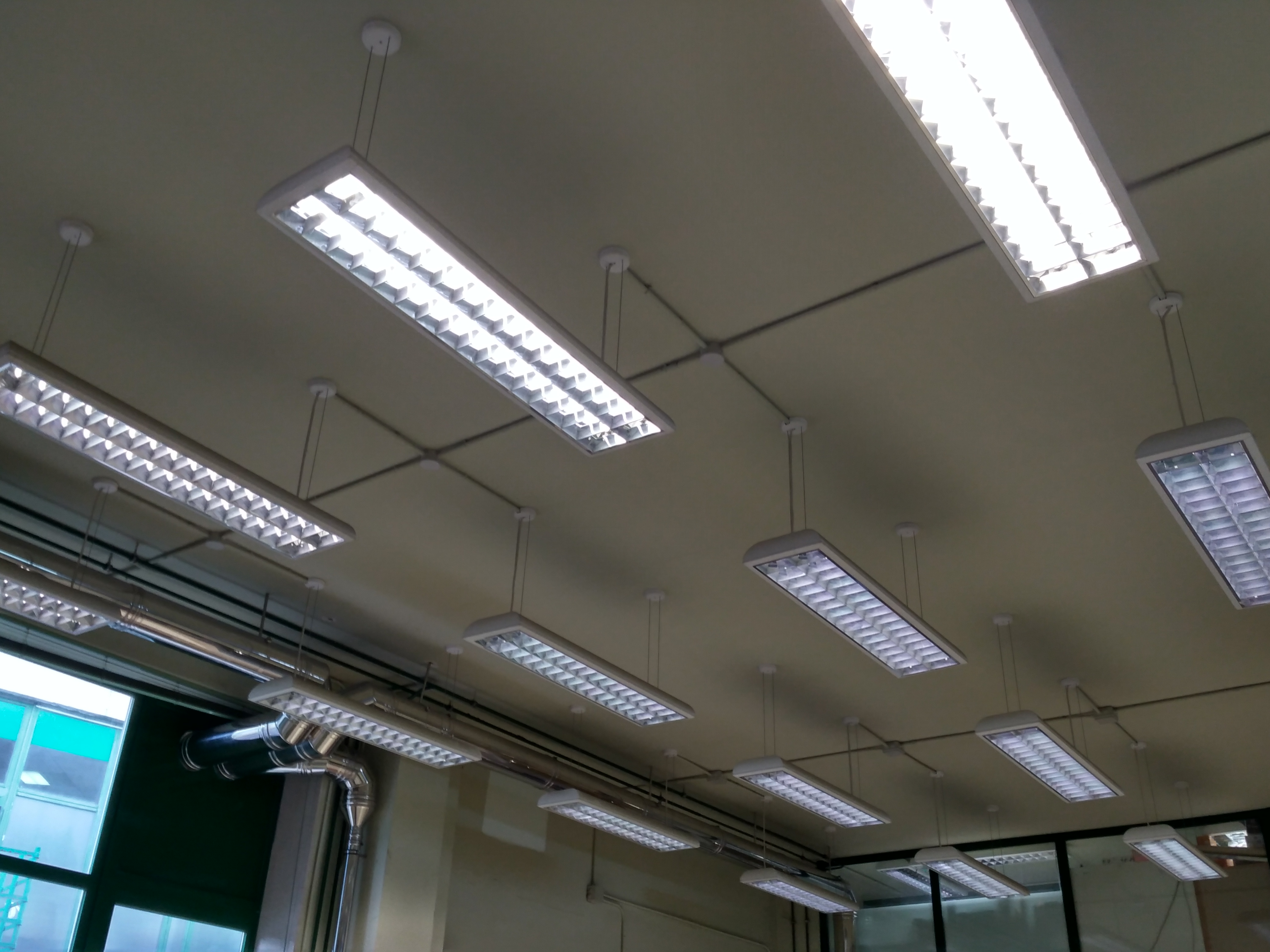}
\caption{}
\end{subfigure}
\begin{subfigure}{0.325\linewidth}
\centering
\includegraphics[width=\textwidth]{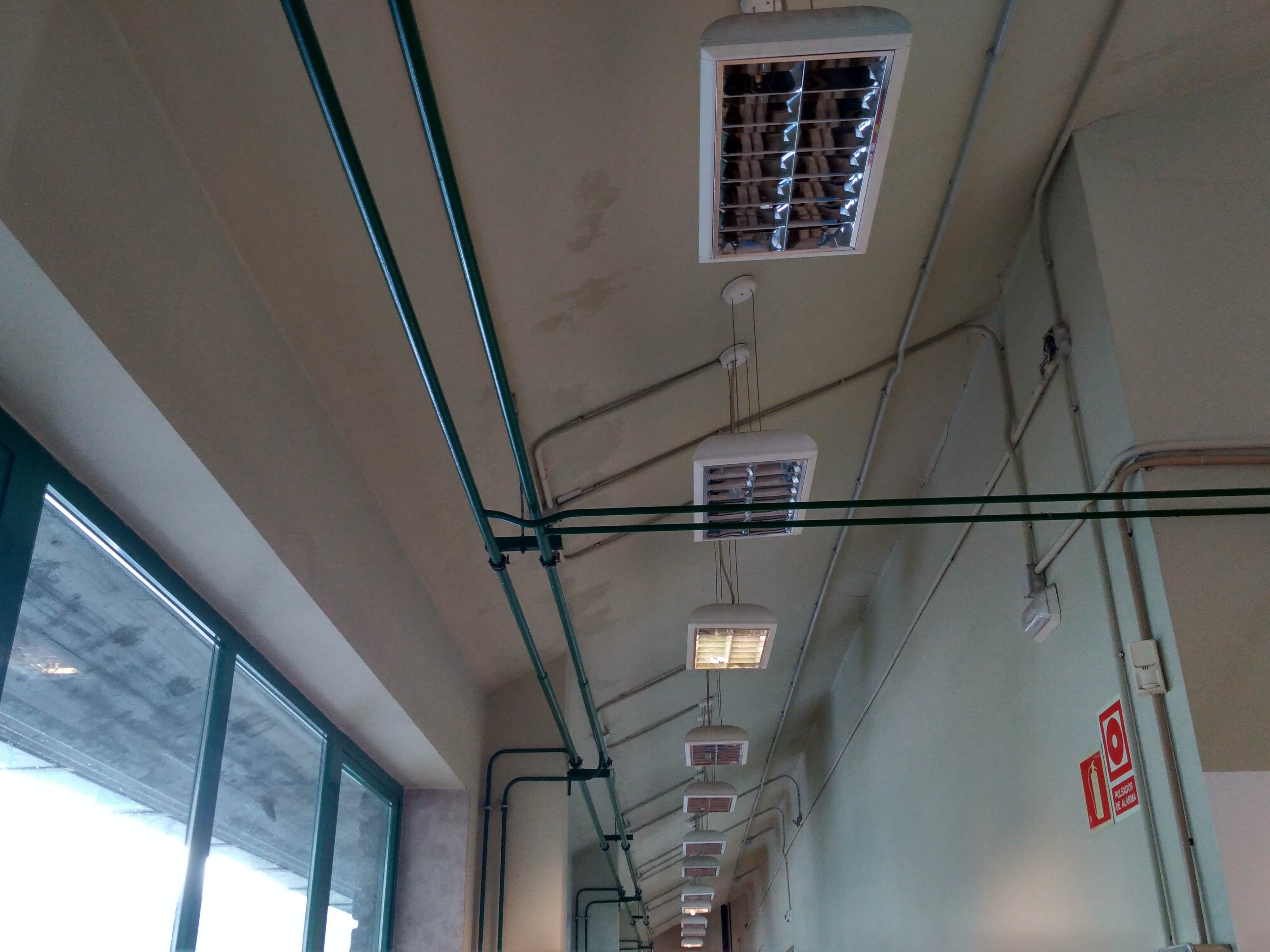}
\caption{}
\end{subfigure}
\begin{subfigure}{0.325\linewidth}
\centering
\includegraphics[width=\textwidth]{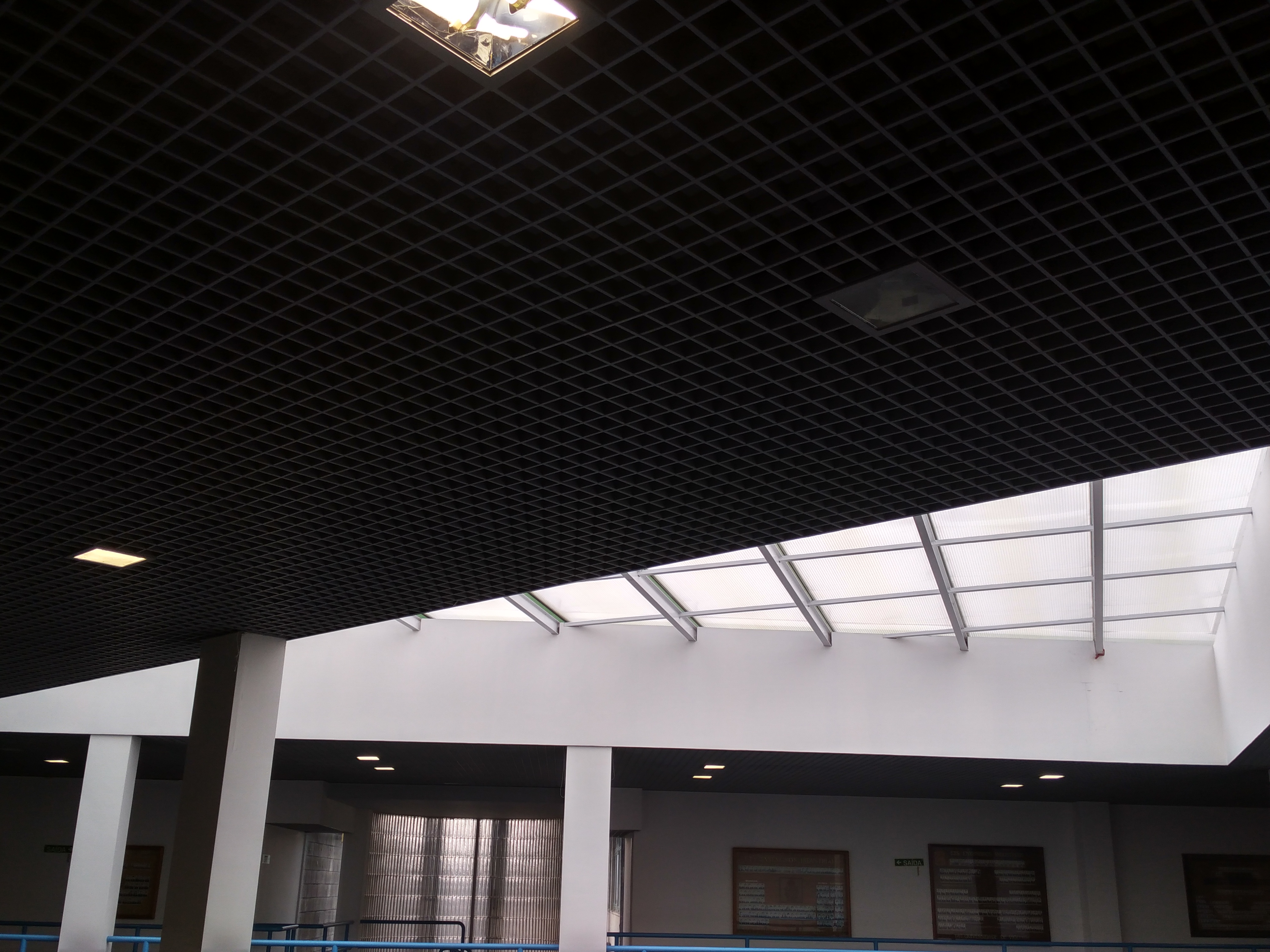}
\caption{}
\end{subfigure}
\begin{subfigure}{0.325\linewidth}
\centering
\includegraphics[width=\textwidth]{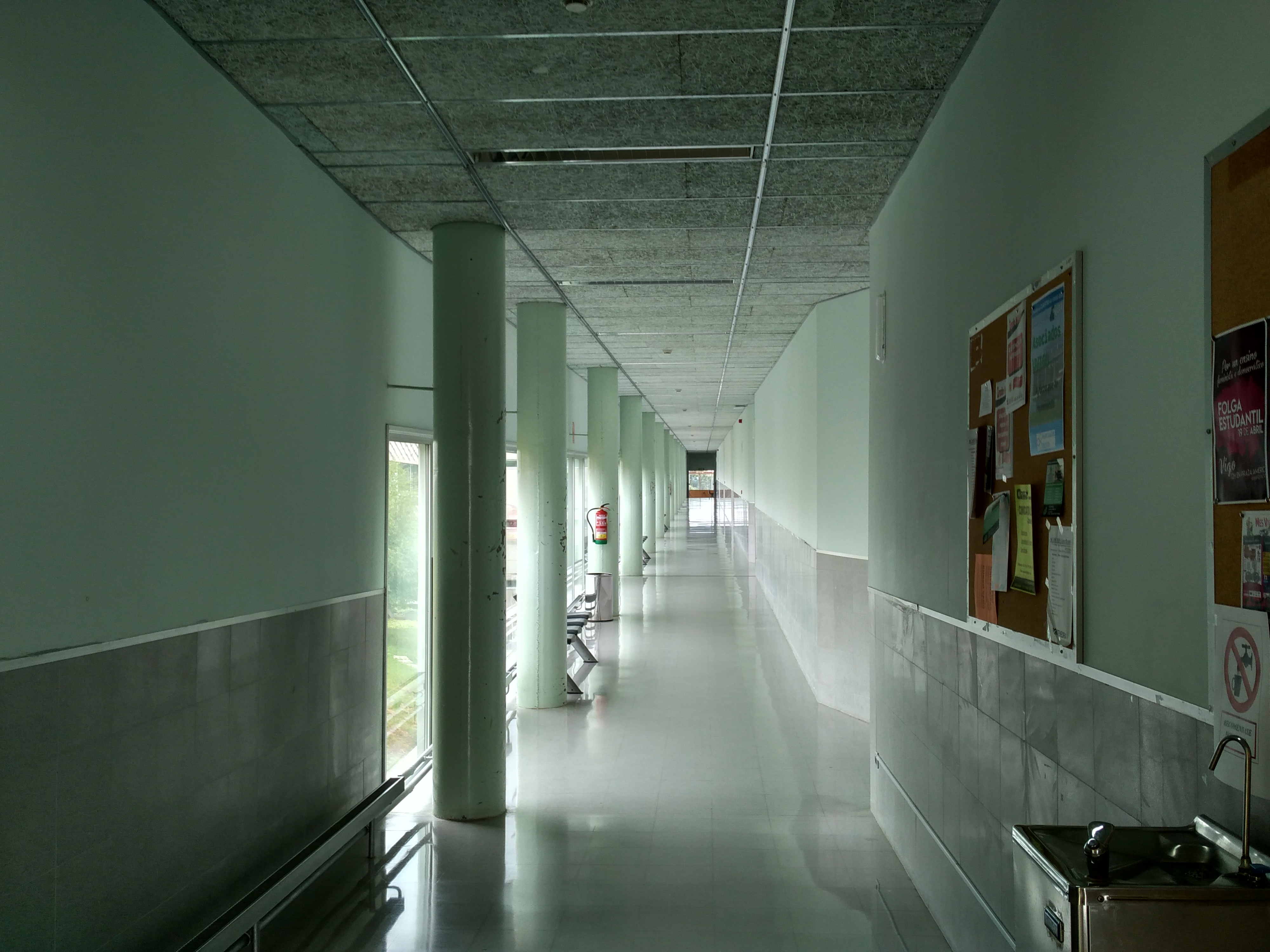}
\caption{}
\end{subfigure}
\begin{subfigure}{0.325\linewidth}
\centering
\includegraphics[width=\textwidth]{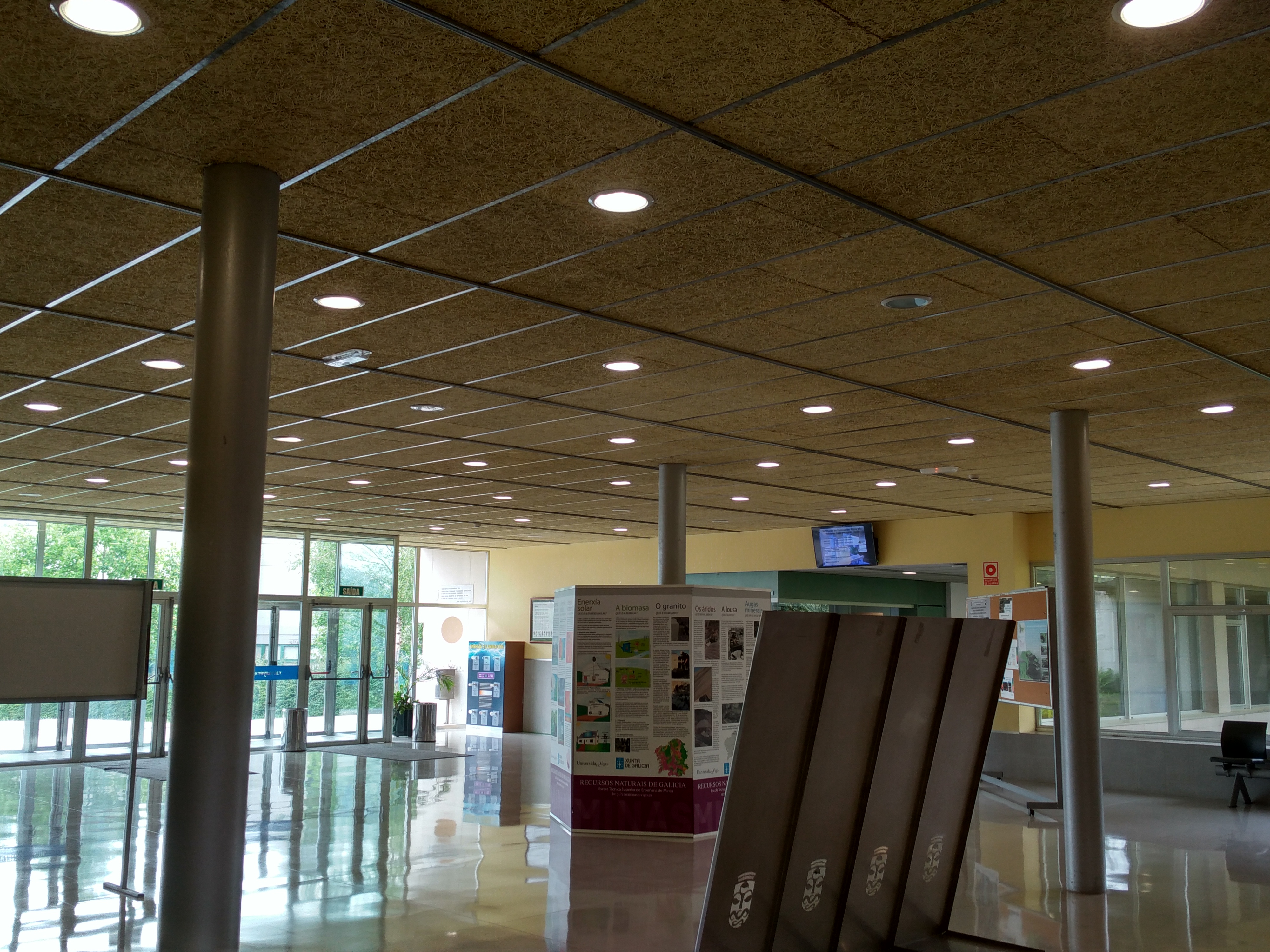}
\caption{}
\end{subfigure}
\caption{Areas with lamps used for the experiments.}
\label{fig:areas}
\end{figure}

\begin{figure}[!ht]
\centering
\includegraphics[width=\linewidth]{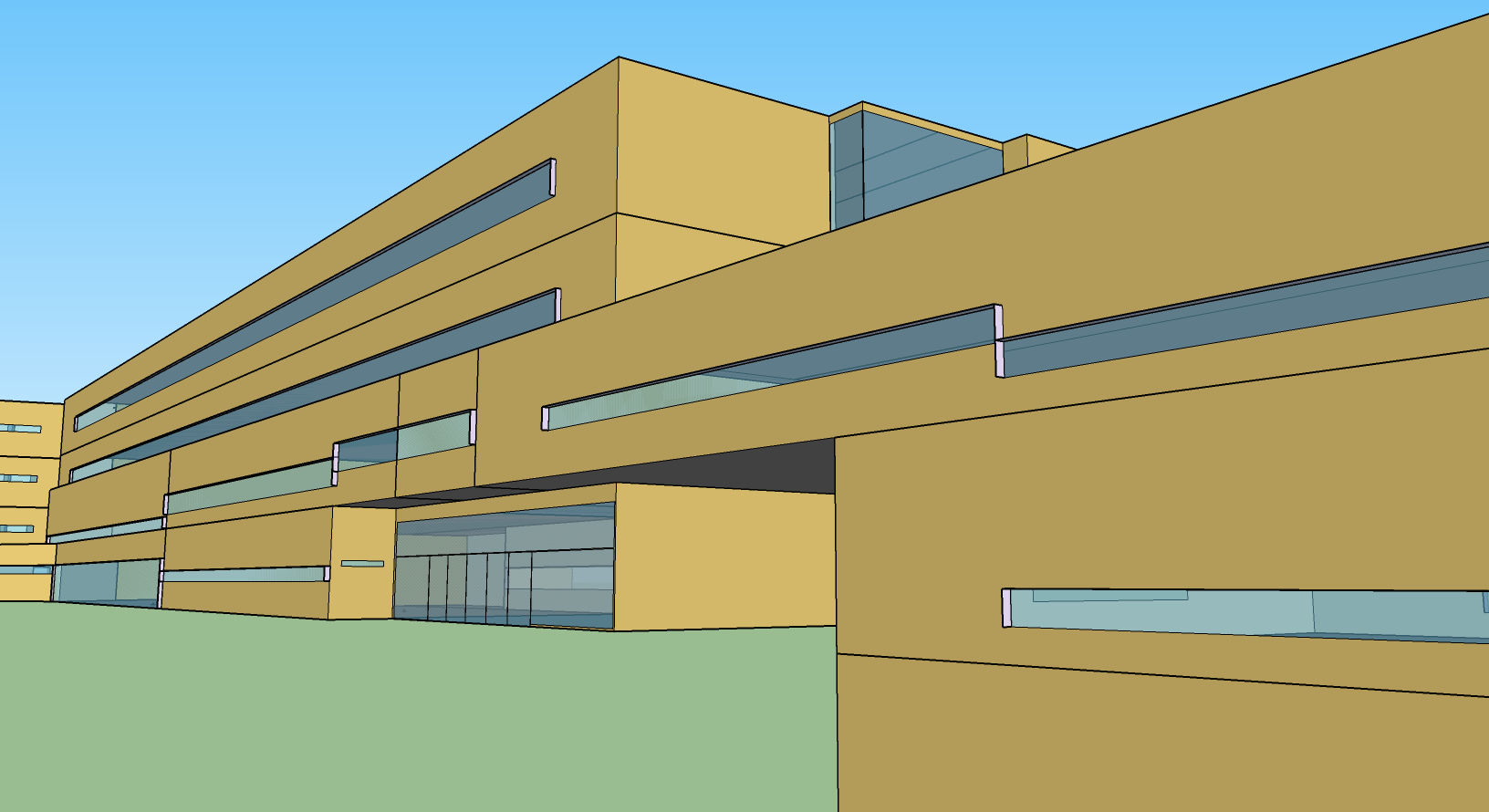}
\caption{Geometry of the BIM model of the building used for the case studies, in SketchUp.}
\label{fig:bim}
\end{figure}

We performed several tests in five case studies with different lamp models. The areas for each case study, displayed in Figure~\ref{fig:areas}, are located in the School of Industrial Engineering and in the School of Mining and Energy Engineering, both in the University of Vigo (Vigo, Spain). The geometry of the BIM model of this building is depicted in Figure~\ref{fig:bim}. The first three areas contain hanging lamps, while the last two contain embedded ones; regarding shape, the first four areas have rectangular lamps, whereas the last one has circular ones. The model database used in the experiments is the same as the one presented in our previous work~\cite{Troncoso2} and the contents of the dataset are included in Table~\ref{tab:inventory}.

The acquisition was performed using a Lenovo Phab 2 Pro with Google Tango~\cite{Tango}, producing grayscale localized images at approximately 30 frames per second, with a resolution of 1920x1080 px, later downscaled to 960x540~px before processing, using a Gaussian pyramid for the downsampling operation~\cite{LearningOpenCV}.

\begin{table*}[!ht]
\centering
\small
\caption{Information of the dataset used for the experiments.}
\begin{tabularx}{\textwidth}{Xcccccc}
\hline
\rotatebox[origin=l]{80}{\textbf{Area}} &
\rotatebox[origin=l]{80}{\textbf{Model}} &
\rotatebox[origin=l]{80}{\textbf{No. images}} &
\rotatebox[origin=l]{80}{\textbf{No. lamps}} &
\rotatebox[origin=l]{80}{\textbf{No. lamps on}} \\
\hline
Laboratory, lamps suspended 50 cm from the ceiling, only two external windows, 1 m from the closest lamps & 1 & 5,674 & 16 & 16 \\
Hallway, lamps suspended 40 cm from the ceiling, external windows at one side & 2 & 2,453 & 19 & 10 \\
Reception, large open area, second floor, lamps fixed at the ceiling, bright environment & 3 & 2,539 & 16 & 13 \\
Hallway, rectangular lamps embedded in the ceiling, external windows at one side & 4 & 6,082 & 25 & 17 \\
Reception, circular lamps embedded in the ceiling & 5 & 14,535 & 90 & 67 \\
\hline
\multicolumn{2}{l}{\textbf{TOTAL}} & 31,283 & 166 & 123 \\
\hline
\end{tabularx}
\label{tab:inventory}
\end{table*}

Reference values were obtained using manual inspection for areas 1-3 and from high-accuracy point cloud information for areas 4 and 5. For area 4, a backpack-based inspection system based on LiDAR sensors and an inertial measurement unit (IMU) was used~\cite{backpack,isprs}, and the point cloud for area 5 was obtained using a FARO Focus3D X 330 Laser Scanner. The details of the two point clouds and measurement systems can be found in~\cite{Troncoso2}.

The methods presented in Sections~\ref{sec:method-opt} and~\ref{sec:method-cls} were developed in C++ with the following supporting software libraries: OpenCV~\cite{OpenCV} for basic artificial vision algorithms, OpenMesh~\cite{OpenMesh} to process 3D information, Ceres Solver~\cite{Ceres} for the base optimization tools, and OpenGL~\cite{OpenGL} for the extraction of visible edge information.

\section{Results}
\label{sec:results}

\begin{figure*}[!ht]
\centering
\begin{subfigure}{0.325\linewidth}
\centering
\includegraphics[width=\textwidth]{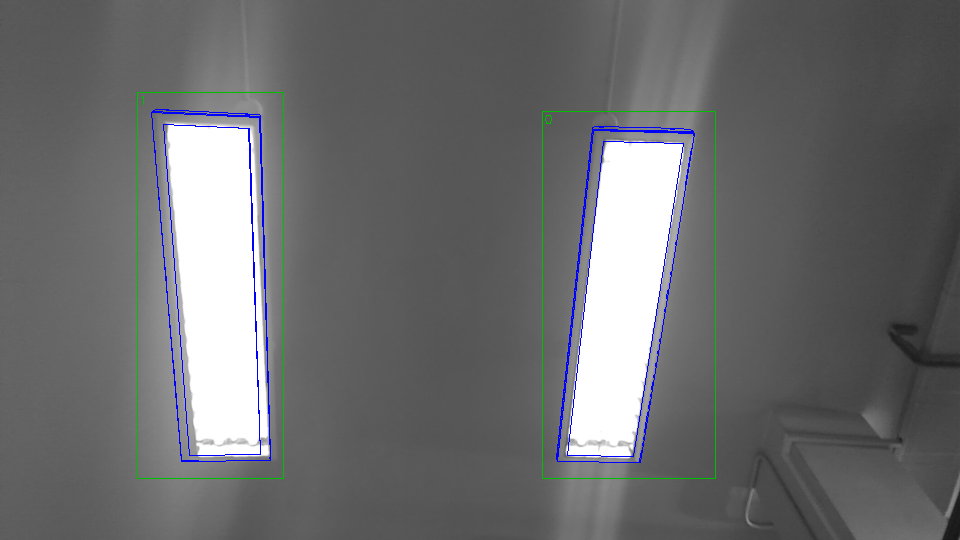}
\caption{}
\end{subfigure}
\begin{subfigure}{0.325\linewidth}
\centering
\includegraphics[width=\textwidth]{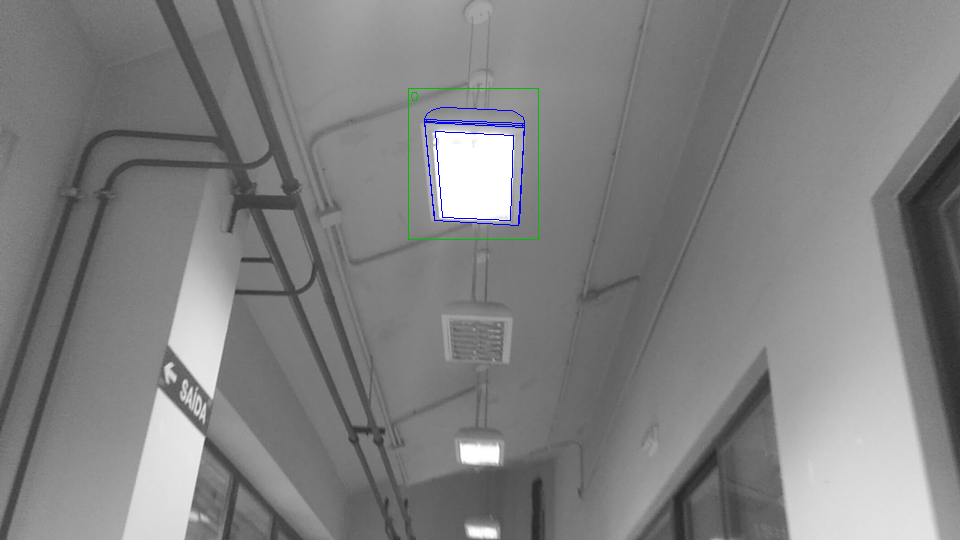}
\caption{}
\end{subfigure}
\begin{subfigure}{0.325\linewidth}
\centering
\includegraphics[width=\textwidth]{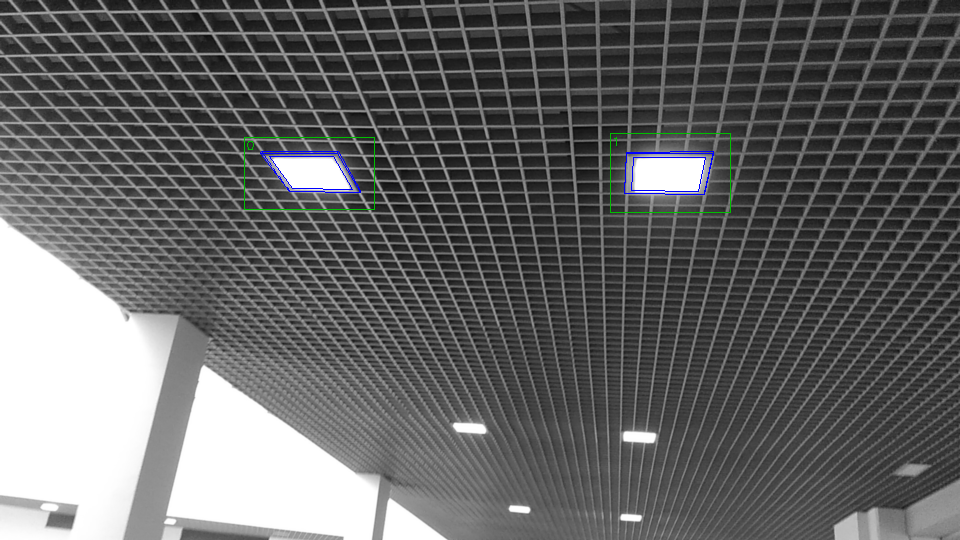}
\caption{}
\end{subfigure}
\begin{subfigure}{0.325\linewidth}
\centering
\includegraphics[width=\textwidth]{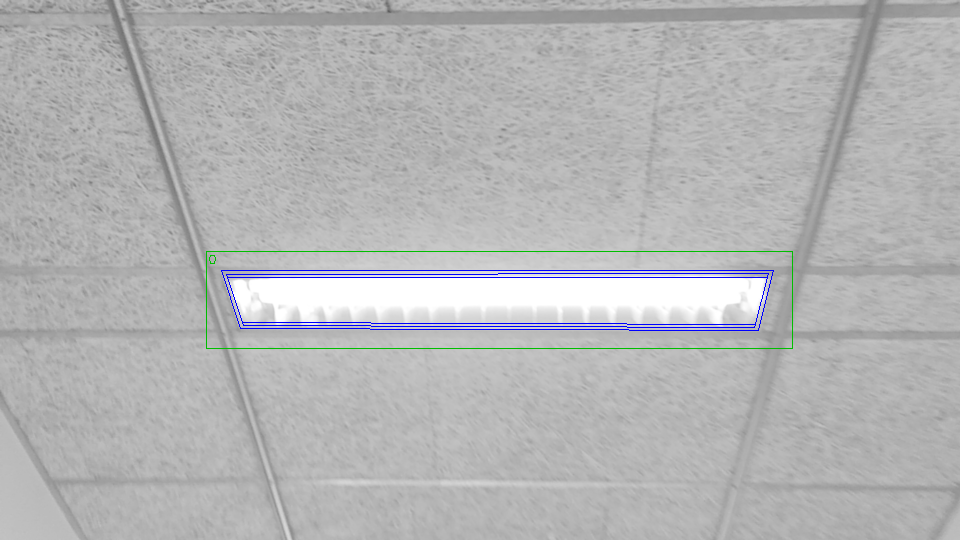}
\caption{}
\end{subfigure}
\begin{subfigure}{0.325\linewidth}
\centering
\includegraphics[width=\textwidth]{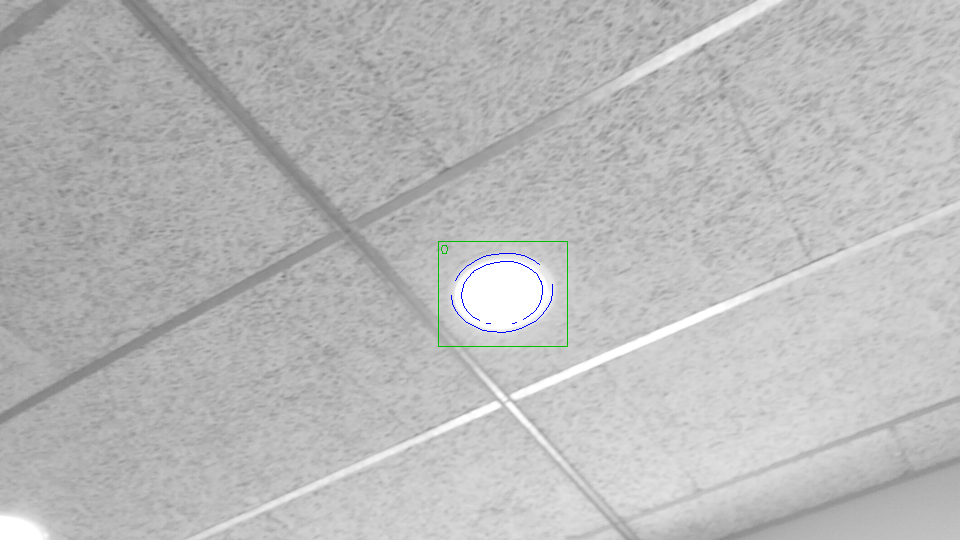}
\caption{}
\end{subfigure}
\caption{Example of detections for each of the lamp models in the database.}
\label{fig:det_imgs}
\end{figure*}

In this section, we present the results for the experiments performed with the goal of evaluating quantitatively the technical contributions presented in previous sections. We include the relevant data obtained from the five case studies described in Section~\ref{sec:exp}, with some example detections for each lamp model displayed in Figure~\ref{fig:det_imgs}.

The general results of the acquisition process are shown in Figures~\ref{fig:dets_1} to \ref{fig:dets_5}, including the individual detections and the final centers after the clustering process, with the reference values for the lamps that are turned on and off. The results displayed in these figures were obtained with the D$^2$CO-IT method with a discretization step (Section~\ref{subsec:error}) of 25\%; this is the configuration used in the rest of the experiments when not otherwise specified.

\newcommand{\detsimgs}[1]{
\begin{figure*}[!ht]
\centering
\begin{subfigure}{0.495\linewidth}
\centering
\includegraphics[width=\textwidth]{img/cl_#1_3_0250_1}
\caption{}
\end{subfigure}
\begin{subfigure}{0.495\linewidth}
\centering
\includegraphics[width=\textwidth]{img/cl_#1_3_0250_2}
\caption{}
\end{subfigure}
\caption{Position of detections, cluster centers and reference values with their corresponding on/off state for model #1: left, without plane estimation; right, with plane estimation and filtering.}
\label{fig:dets_#1}
\end{figure*}
}

\detsimgs{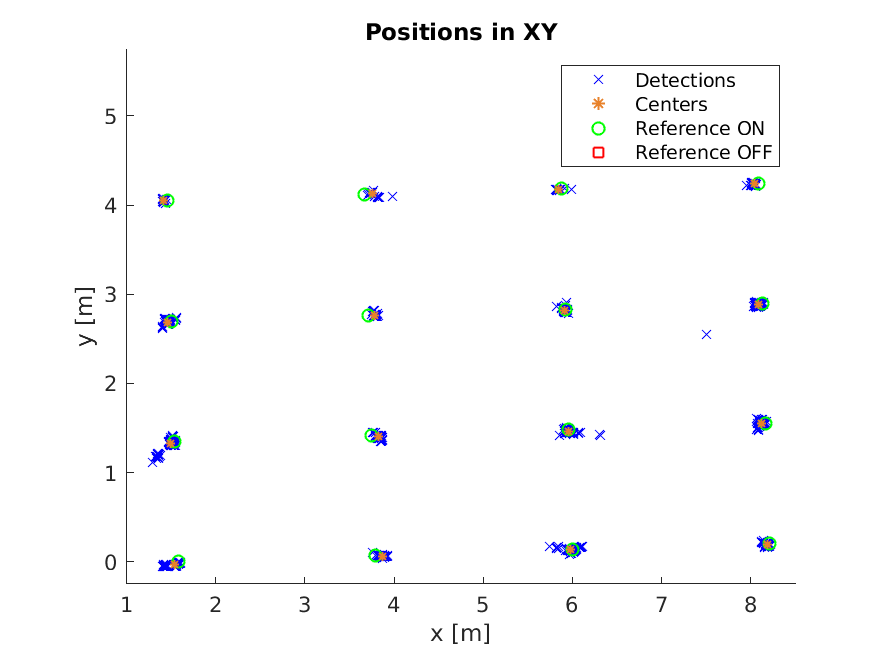}
\detsimgs{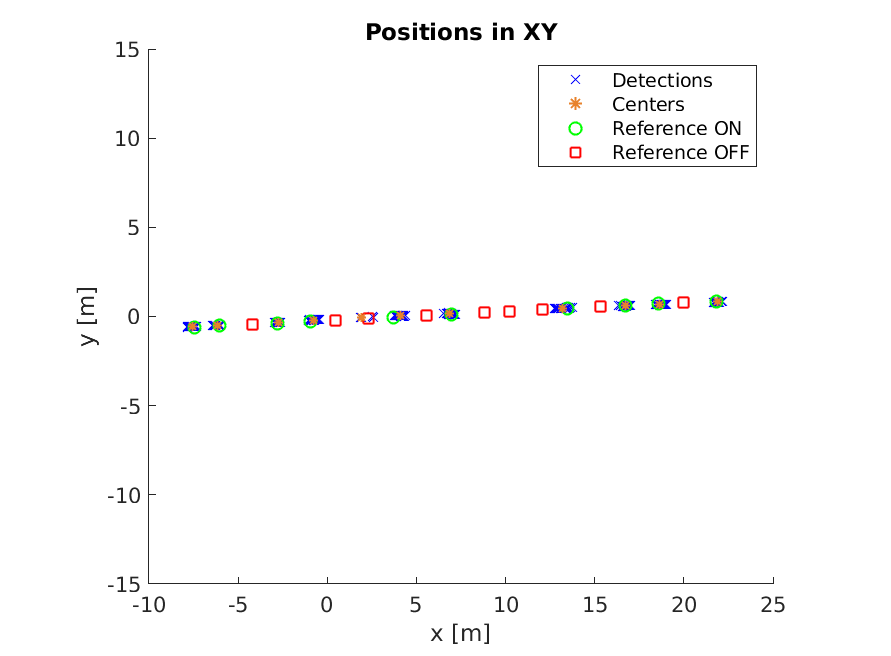}
\detsimgs{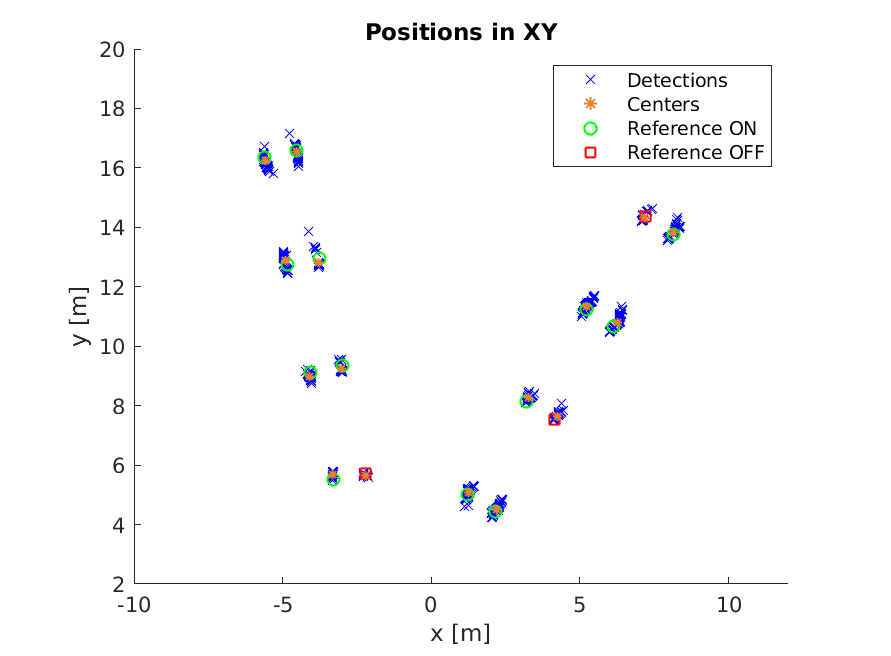}
\detsimgs{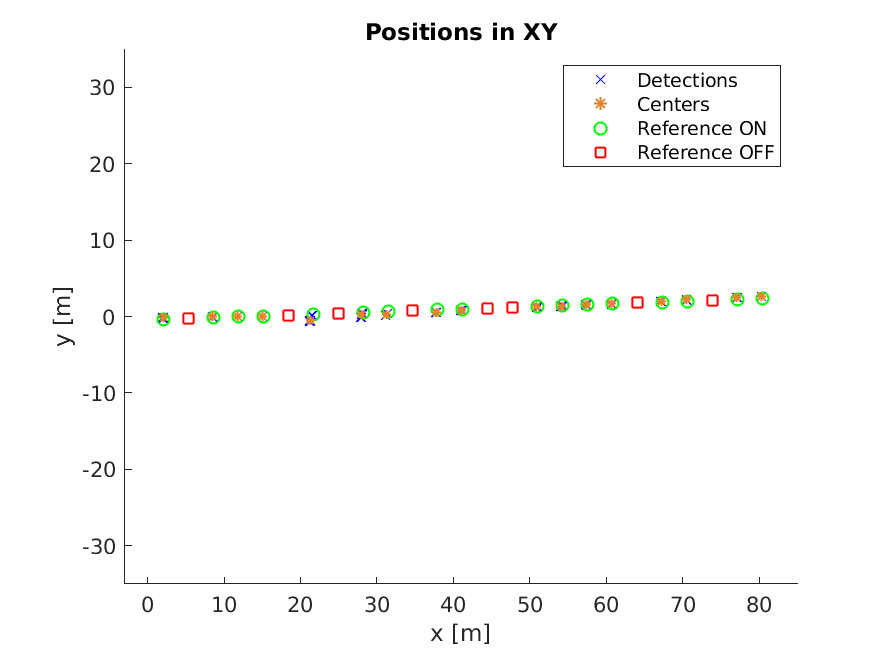}
\detsimgs{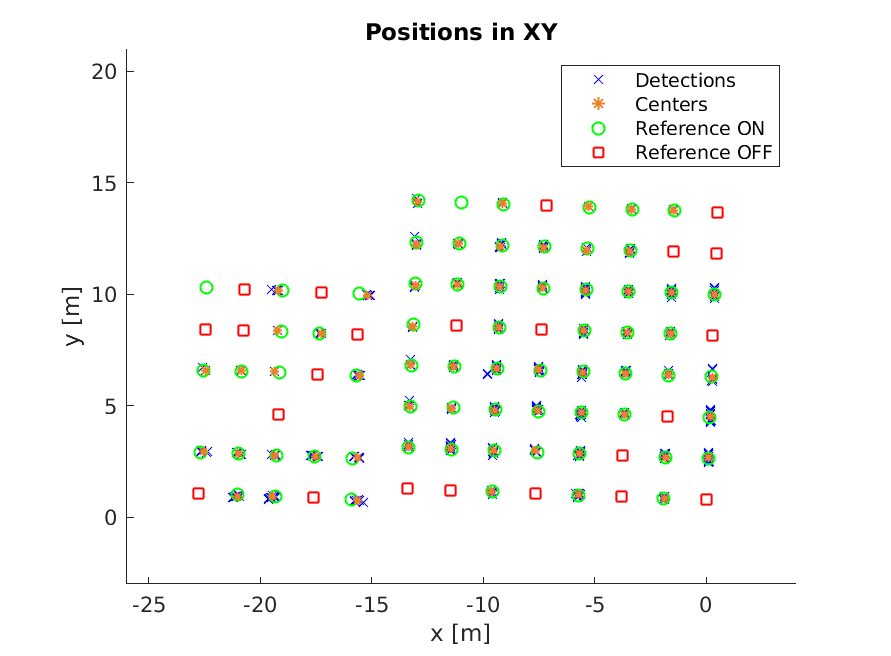}

We analyze the performance in three main categories: \emph{detection}, \emph{identification} and \emph{localization}. For the detection, we include figures for the number of times a valid element is detected (Section~\ref{subsec:res-det}); in the identification (Section~\ref{subsec:res-idt}), we examine which lamp model is detected each time with respect to the corresponding correct model and what state is perceived; and finally, we evaluate the localization performance (Section~\ref{subsec:res-loc}) in terms of the distance between the positions of the detected and real lamps and between the positions of the detections inside each cluster. The results for the three categories incorporate comparisons between different optimization methods and plane projection strategies.

\subsection{Detection}
\label{subsec:res-det}

First, we evaluate the performance of the system in terms of the detection rate. We compare this parameter for different optimization methods, including the original D$^2$CO~\cite{D2CO}, our proposed improvement, D$^2$CO-IT, and an intermediate version of the method that uses the values of all the pixels for each edge but with the DT3$_V$ instead of the IDT3$_V$. We call this method D$^2$CO-E.

\begin{figure}[!ht]
\centering
\begin{subfigure}{0.48\textwidth}
\centering
\includegraphics[width=\textwidth]{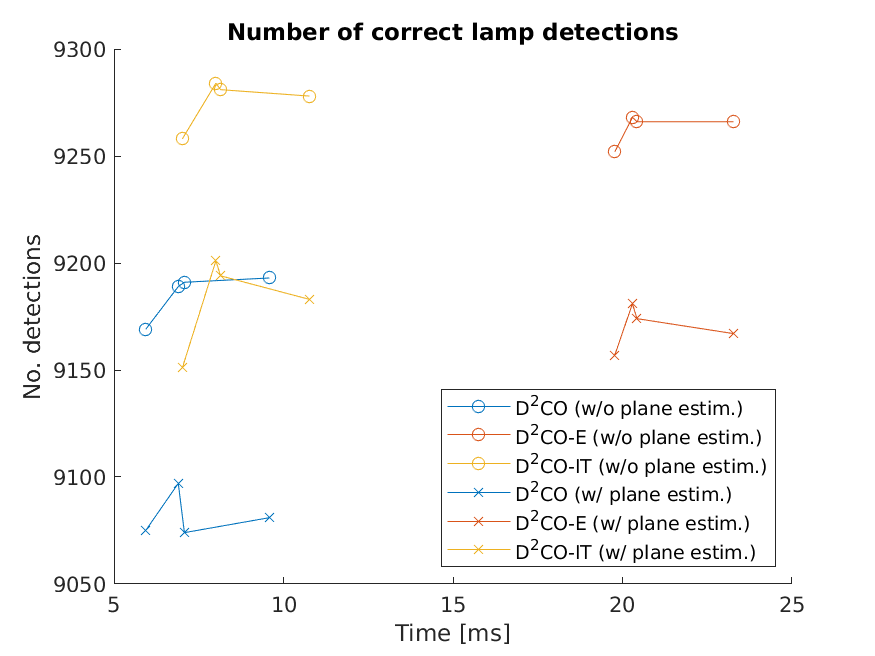}
\caption{}
\label{fig:times_dets}
\end{subfigure}
\begin{subfigure}{0.48\textwidth}
\centering
\includegraphics[width=\textwidth]{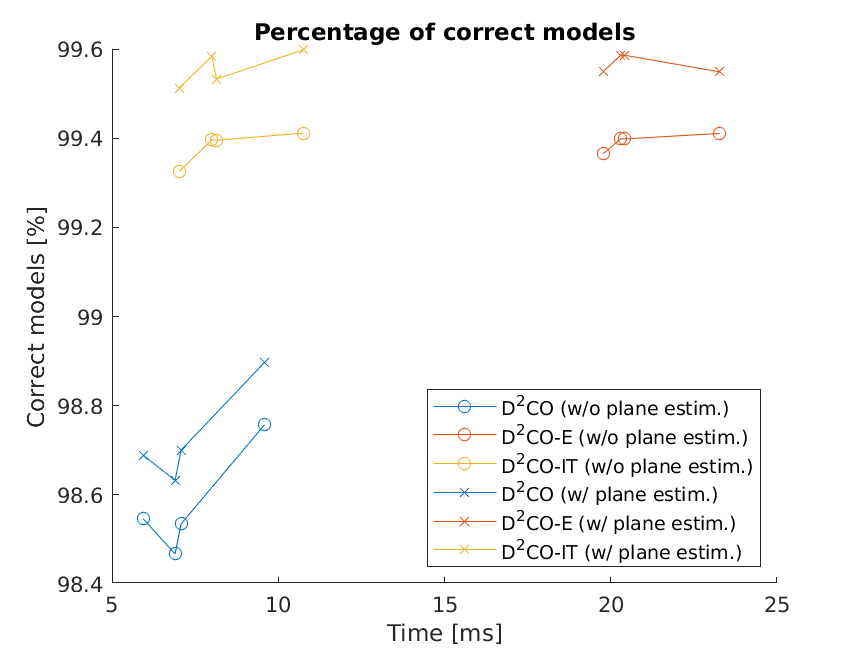}
\caption{}
\label{fig:times_mods}
\end{subfigure}
\caption{Detection and identification statistics with and without plane estimation and filtering for different optimization methods with step values of 25\%, 50\%, 75\% and 100\% of the largest edge in the model.}
\label{fig:times_dets_mods}
\end{figure}

The results for different combinations of optimization method and discretization step are shown in Figure~\ref{fig:times_dets}. We can see that the use of all the pixel values improves the detection rate for both D$^2$CO-E and D$^2$CO-IT, with and without plane estimation and filtering.
Moreover, the use of the integral tensor in D$^2$CO-IT greatly improves the speed of the method compared to D$^2$CO-E, resulting in times comparable to D$^2$CO, even when all the pixel values are used. The average values for all the steps are shown in Table~\ref{tab:times_dets}.

\begin{table}[!ht]
\centering
\small
\caption{Average time and number of correct lamp detections over step values of 25\%, 50\%, 75\% and 100\% for different optimization methods with (w/) and without (w/o) plane estimation and filtering.}
\begin{tabular}{lccc}
\hline
 & \textbf{D$^2$CO} & \textbf{D$^2$CO-E} & \textbf{D$^2$CO-IT} \\
\hline
Time [ms] & 8.2387 & 21.8801 & 9.4107 \\
No. dets. w/o p. e. & 9,198.1 & 9,264.7 & 9,277.5 \\
No. dets. w/  p. e. & 9,096.5 & 9,172.9 & 9,189.6 \\
\hline
\end{tabular}
\label{tab:times_dets}
\end{table}

Finally, the use of plane estimation decreases the number of final detections because there is a filtering step excluding those too distant from the expected plane. Although there is a lesser number of detections, the overall quality of those is higher, as will be shown in the identification and localization analyses.
Additionally, Table~\ref{tab:stats_base} includes the detection rates for the D$^2$CO-IT method with a discretization step of 25\% for each of the five lamp models.

\begin{table}[!ht]
\centering
\small
\caption{Number of correct lamp detections for the D$^2$CO-IT with a step of 25\%.}
\begin{tabular}{ccc}
\hline
\textbf{Model} & \textbf{Without plane estim.} & \textbf{With plane estim.} \\
\hline
1 & 1,973 & 1,968 \\
2 & 832 & 783 \\
3 & 810 & 801 \\
4 & 507 & 507 \\
5 & 5,156 & 5,124 \\
\hline
TOTAL & 9,278 & 9,183 \\
\hline
\end{tabular}
\label{tab:stats_base}
\end{table}

\subsection{Identification}
\label{subsec:res-idt}

The second analysis corresponds to the identification of both the correct lamp type and the correct state of the lamps. Figures~\ref{fig:distrib_1} and~\ref{fig:distrib_2} show the distribution of accumulated scores of the individual detections for each cluster for the five distinct lamp types in each area of study. While some of the detections are not correct, the maximum score matches the appropriate lamp model for 100\% of the 65 clusters, independent of the plane estimation step.

\newcommand{\distribimgs}[2]{
\begin{figure*}[!ht]
\centering
\begin{subfigure}{0.325\linewidth}
\centering
\includegraphics[width=\textwidth]{img/distrib_1_#1}
\caption{}
\end{subfigure}
\begin{subfigure}{0.325\linewidth}
\centering
\includegraphics[width=\textwidth]{img/distrib_2_#1}
\caption{}
\end{subfigure}
\begin{subfigure}{0.325\linewidth}
\centering
\includegraphics[width=\textwidth]{img/distrib_3_#1}
\caption{}
\end{subfigure}
\begin{subfigure}{0.325\linewidth}
\centering
\includegraphics[width=\textwidth]{img/distrib_4_#1}
\caption{}
\end{subfigure}
\begin{subfigure}{0.665\linewidth}
\centering
\includegraphics[width=\textwidth]{img/distrib_5_#1}
\caption{}
\end{subfigure}
\caption{Distribution of accumulated scores for each cluster of the five case studies #2 plane estimation and filtering. Each value on the x-axis contains the sum of the scores for all the detections for each lamp model, with the model corresponding to the highest accumulated value included below.}
\label{fig:distrib_#1}
\end{figure*}
}

\distribimgs{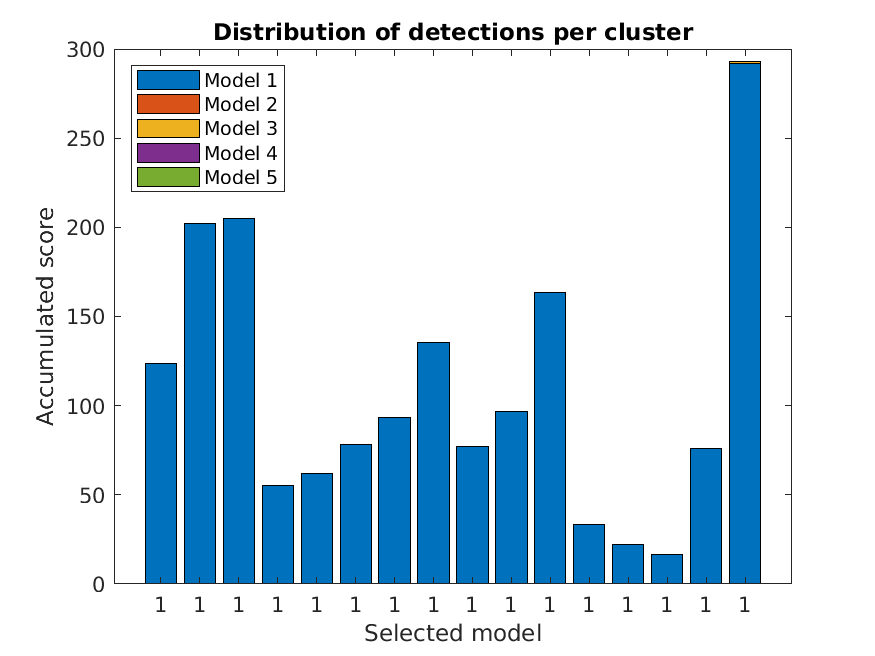}{without}
\distribimgs{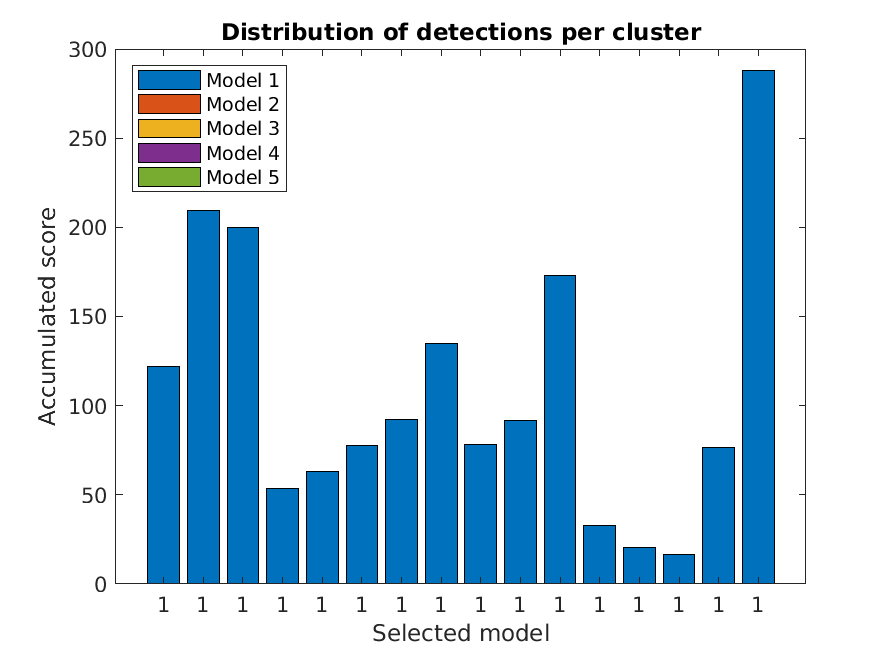}{with}

The great majority of errors correspond to mismatches between Model~3 and Model~5. This is due to both of them having shapes in the image with similar circularity~\cite{Rosin}, which is used to differentiate between circular and polygonal lamps~\cite{Troncoso2}, especially with poor lighting conditions that result in glows in the image.

\begin{figure}[!ht]
\centering
\begin{subfigure}{0.44\textwidth}
\centering
\includegraphics[width=\textwidth]{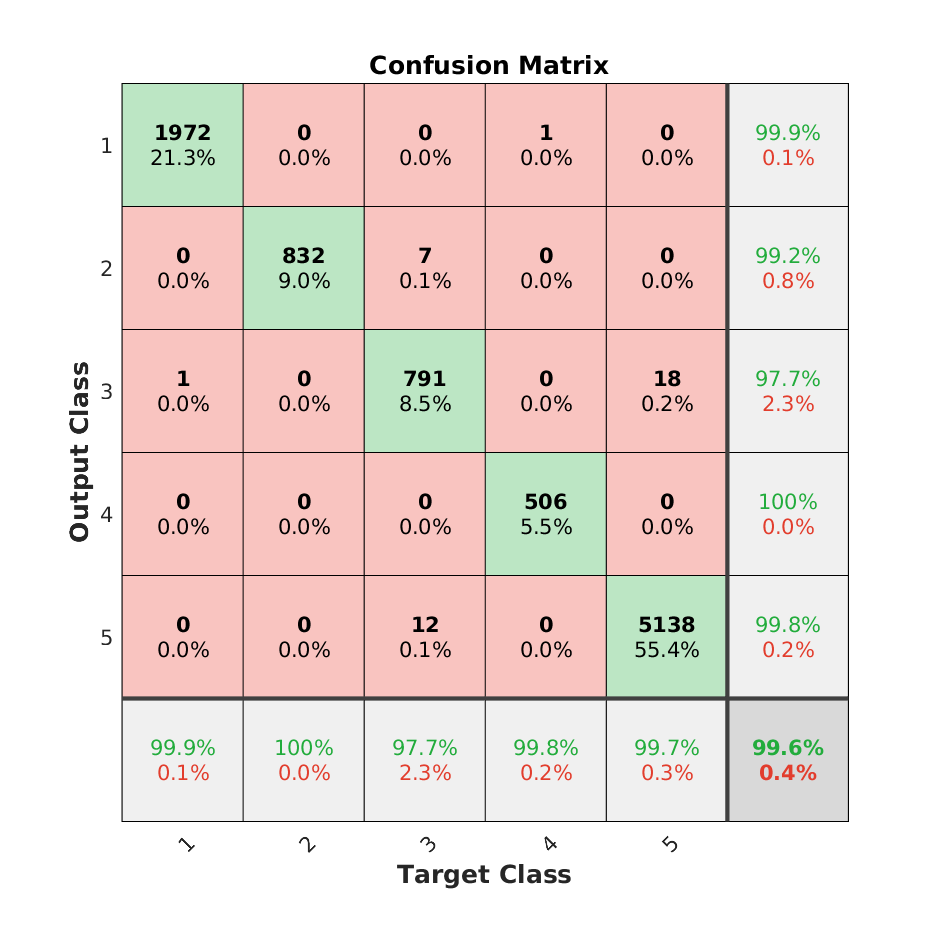}
\caption{}
\end{subfigure}
\begin{subfigure}{0.44\textwidth}
\centering
\includegraphics[width=\textwidth]{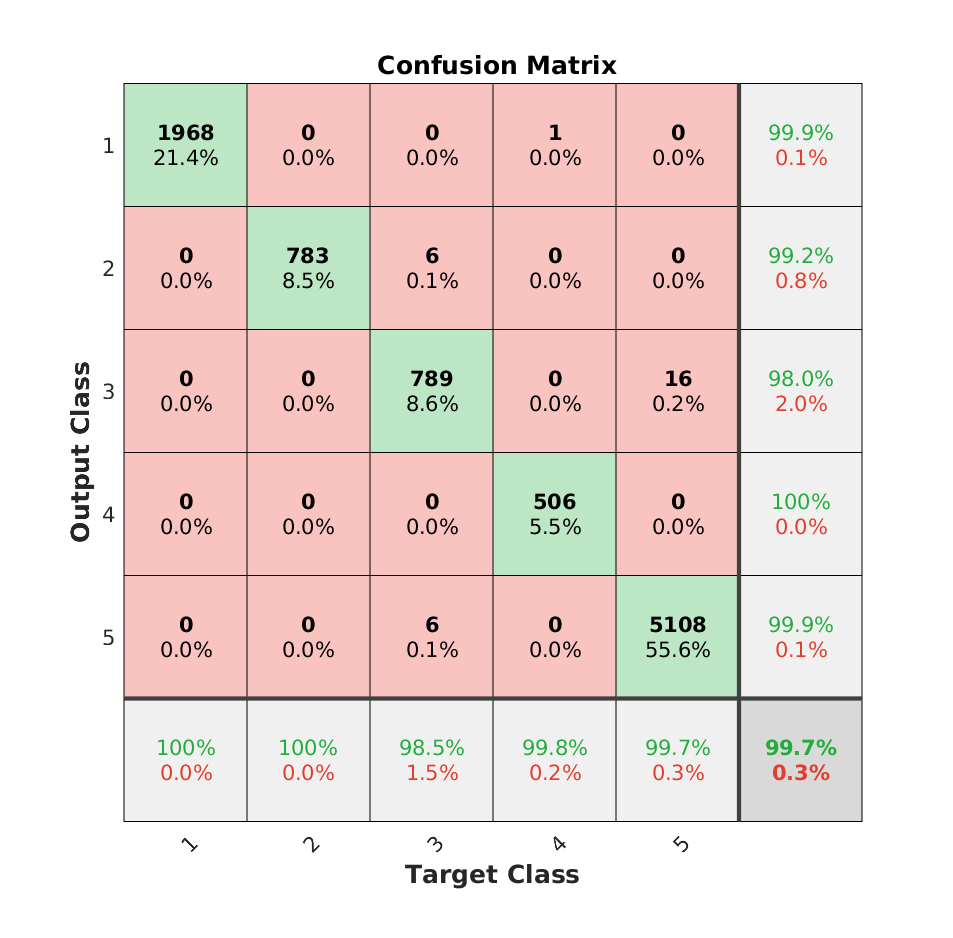}
\caption{}
\end{subfigure}
\caption{Confusion matrices for the lamp models used in the experiments. (a) Results without plane estimation and (b) results with plane estimation and filtering.}
\label{fig:conf}
\end{figure}

The confusion matrices, which indicate the number of detections for each of the expected lamp models for the complete set, are displayed in Figure~\ref{fig:conf}. In this case, the use of the plane estimation and filtering decreases the percentage of incorrect identifications from 0.42\% to 0.32\%. The use of plane estimation does not affect the performance of the lamp state identification, which achieves a 97.9\% of correct state detection ratio.

Finally, the optimization method also has an impact on the percentage of correct identifications, as shown in Figure~\ref{fig:times_mods}, with the average values presented in Table~\ref{tab:times_mods}. Here, the use of all the pixel values of the edges lowers the average identification error from 1.37\% to 0.61\% without plane estimation and from 1.23\% to 0.43\% with plane estimation and filtering.

\begin{table}[!ht]
\centering
\small
\caption{Average time and percentage of correct lamp model identifications over step values of 25\%, 50\%, 75\% and 100\% for different optimization methods with (w/) and without (w/o) plane estimation and filtering.}
\begin{tabular}{lccc}
\hline
 & \textbf{D$^2$CO} & \textbf{D$^2$CO-E} & \textbf{D$^2$CO-IT} \\
\hline
Time [ms] & 8.2387 & 21.8801 & 9.4107 \\
Correct ident. w/o p. e. [\%] & 98.6288 & 99.3989 & 99.3927 \\
Correct ident. w/  p. e. [\%] & 98.7720 & 99.5736 & 99.5673 \\
\hline
\end{tabular}
\label{tab:times_mods}
\end{table}

\subsection{Localization}
\label{subsec:res-loc}

\begin{figure*}[!ht]
\centering
\begin{subfigure}{0.3145\linewidth}
\centering
\includegraphics[width=\textwidth]{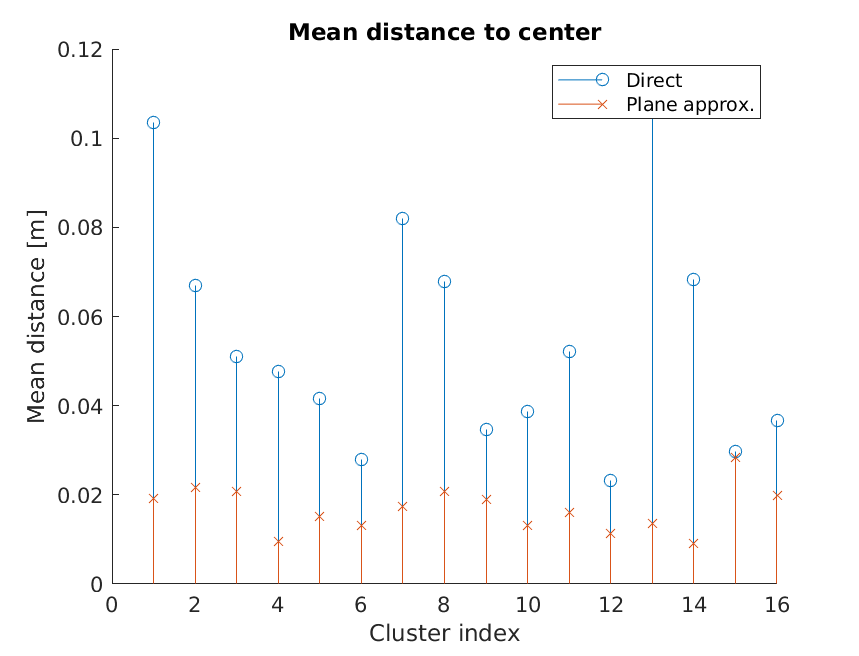}
\caption{}
\end{subfigure}
\begin{subfigure}{0.3145\linewidth}
\centering
\includegraphics[width=\textwidth]{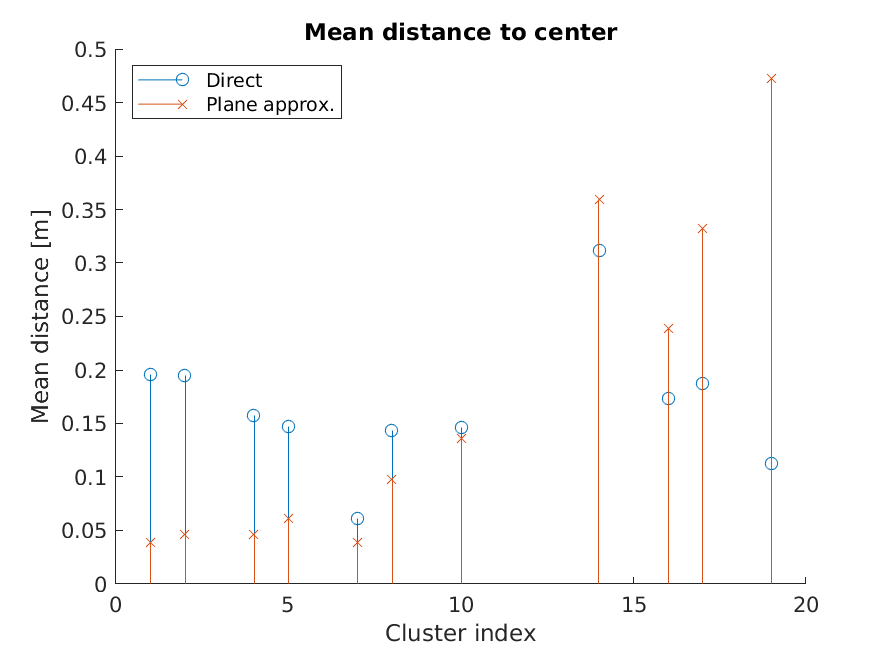}
\caption{}
\end{subfigure}
\begin{subfigure}{0.3145\linewidth}
\centering
\includegraphics[width=\textwidth]{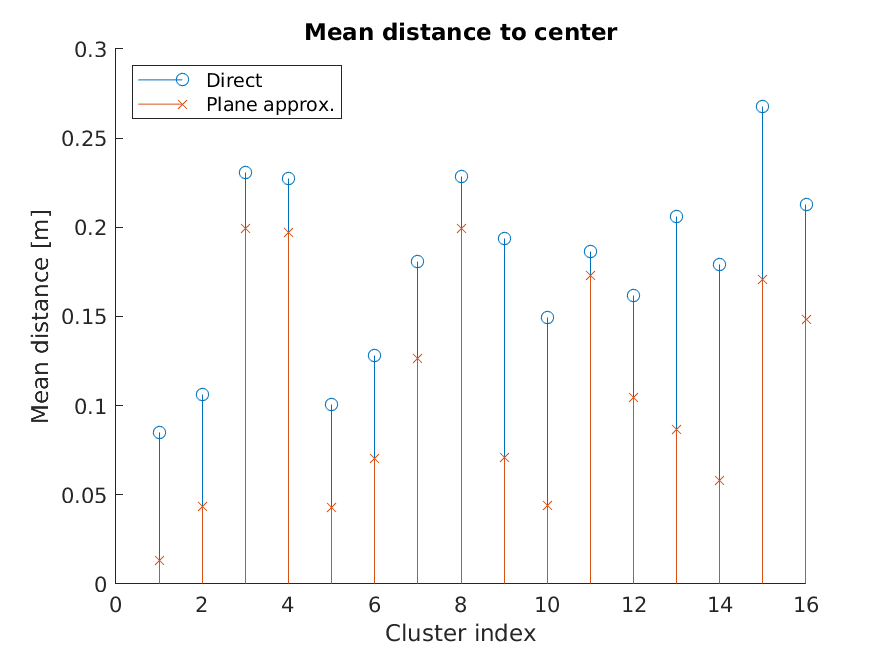}
\caption{}
\end{subfigure}
\begin{subfigure}{0.3145\linewidth}
\centering
\includegraphics[width=\textwidth]{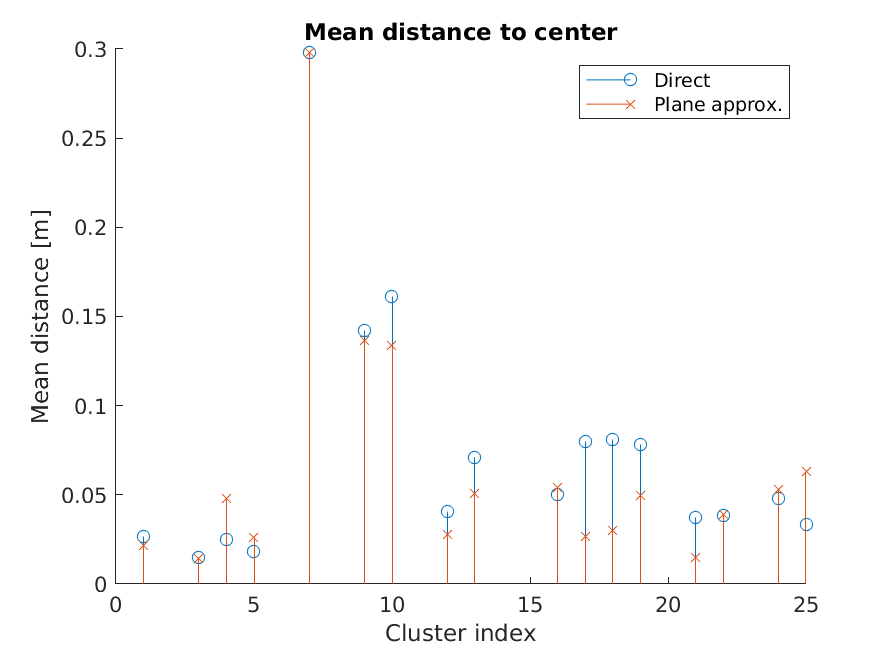}
\caption{}
\end{subfigure}
\begin{subfigure}{0.3145\linewidth}
\centering
\includegraphics[width=\textwidth]{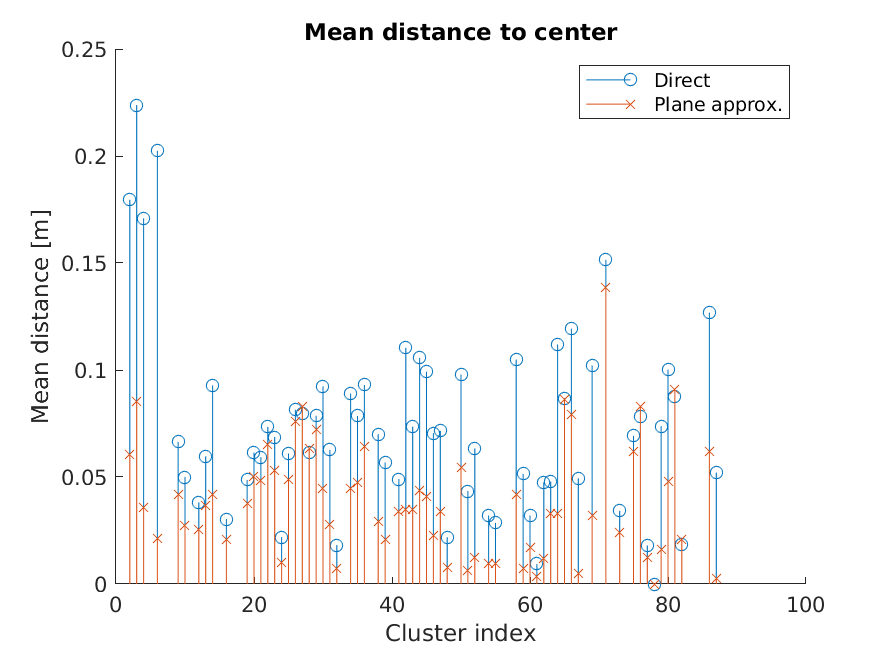}
\caption{}
\end{subfigure}
\begin{subfigure}{0.3145\linewidth}
\centering
\includegraphics[width=\textwidth]{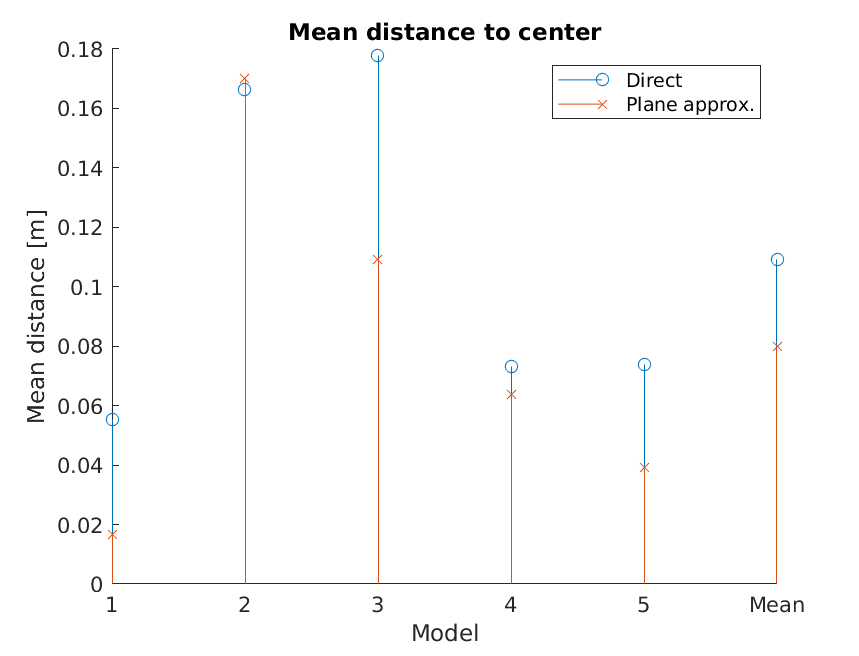}
\caption{}
\end{subfigure}
\caption{Means of individual detection distances to cluster center with and without plane estimation and filtering. (a-e) Models 1-5; (f) global statistics.}
\label{fig:means}
\end{figure*}

\begin{figure*}[!ht]
\centering
\begin{subfigure}{0.3145\linewidth}
\centering
\includegraphics[width=\textwidth]{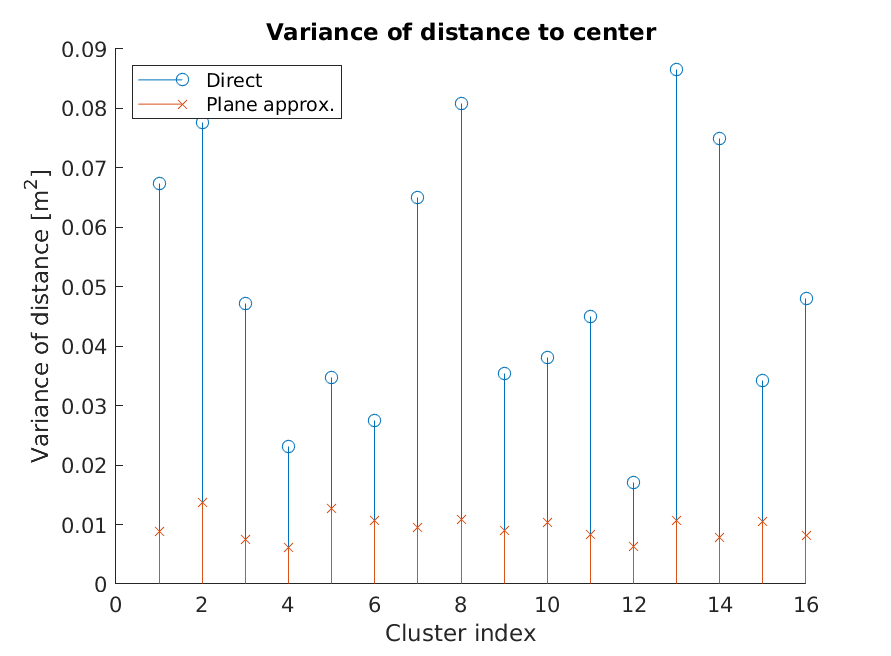}
\caption{}
\end{subfigure}
\begin{subfigure}{0.3145\linewidth}
\centering
\includegraphics[width=\textwidth]{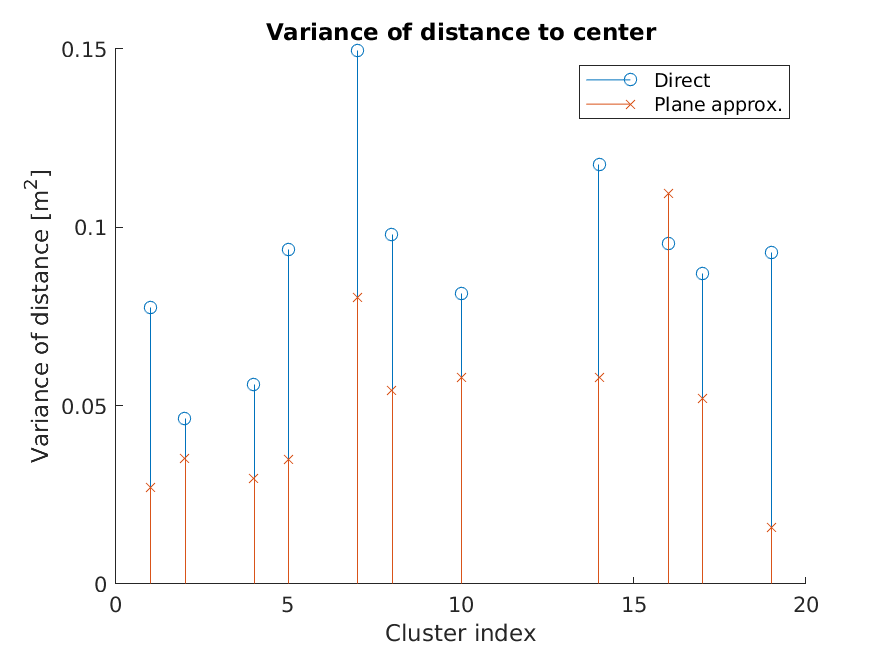}
\caption{}
\end{subfigure}
\begin{subfigure}{0.3145\linewidth}
\centering
\includegraphics[width=\textwidth]{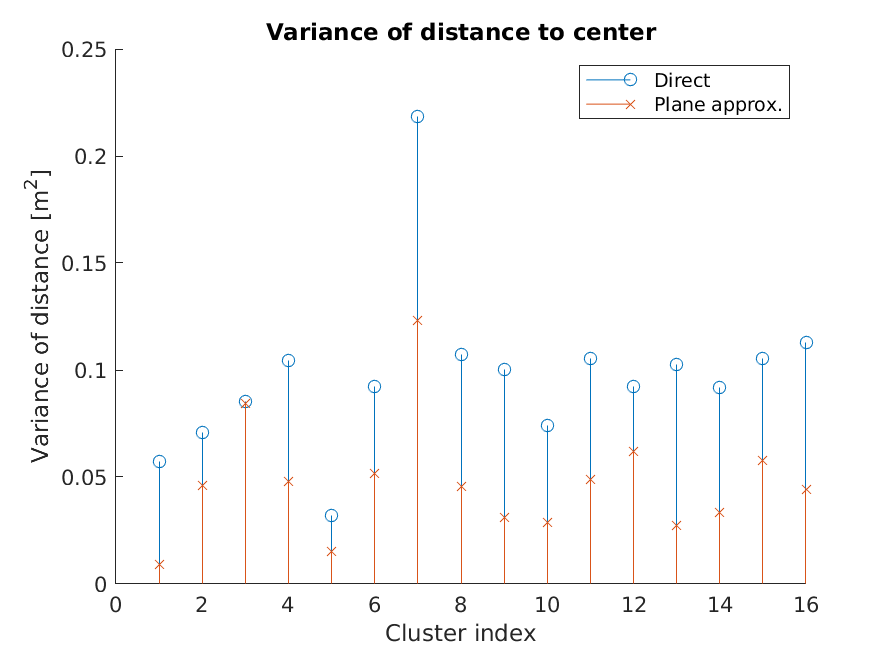}
\caption{}
\end{subfigure}
\begin{subfigure}{0.3145\linewidth}
\centering
\includegraphics[width=\textwidth]{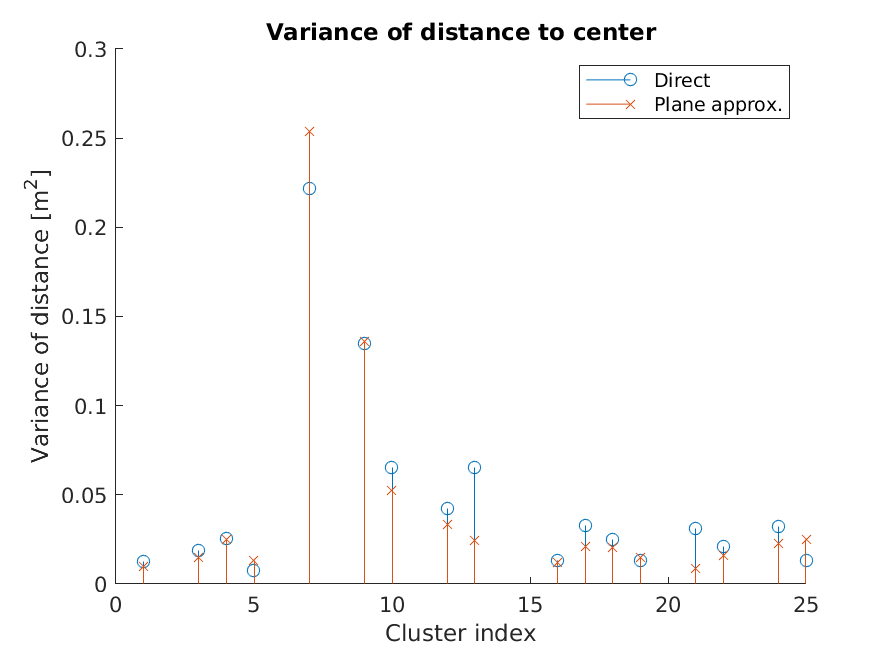}
\caption{}
\end{subfigure}
\begin{subfigure}{0.3145\linewidth}
\centering
\includegraphics[width=\textwidth]{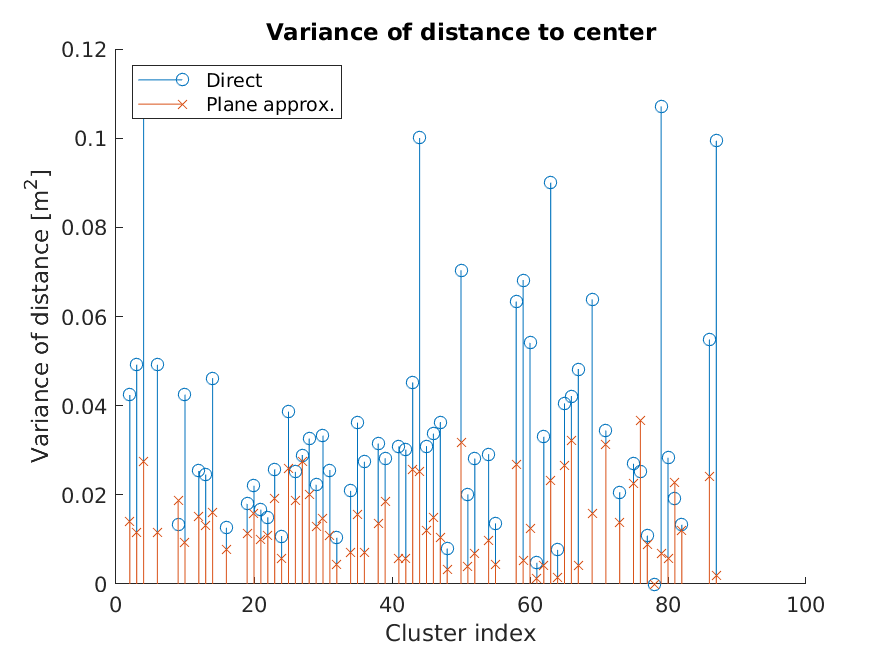}
\caption{}
\end{subfigure}
\begin{subfigure}{0.3145\linewidth}
\centering
\includegraphics[width=\textwidth]{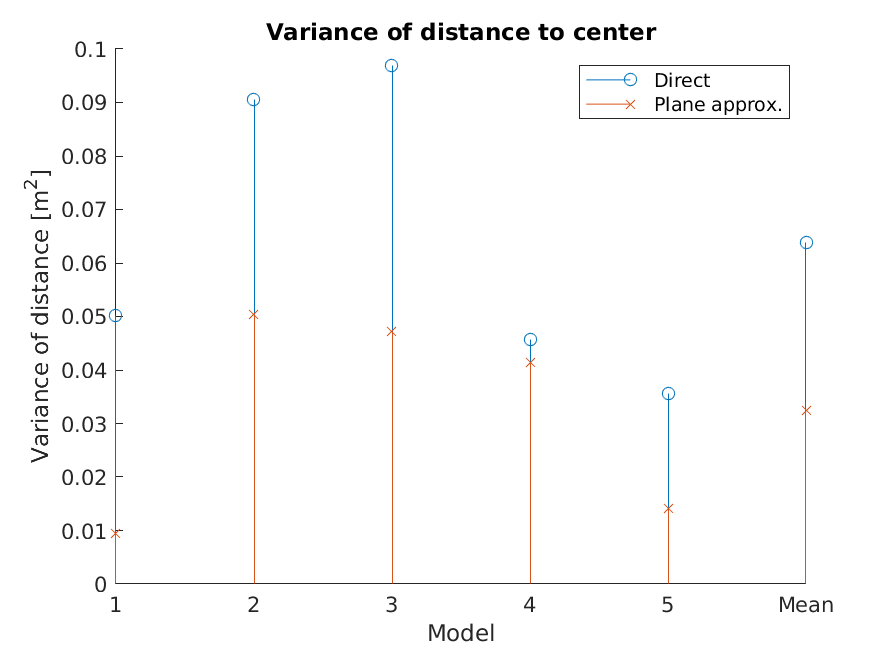}
\caption{}
\end{subfigure}
\caption{Variances of individual detection distances to cluster center with and without plane estimation and filtering. (a-e) Models 1-5; (f) global statistics.}
\label{fig:vars}
\end{figure*}

The last of the analyses covers the localization of the detections. We can see in Figures~\ref{fig:dets_1} to \ref{fig:dets_5} that the use of the plane estimation clearly reduces the dispersion of the detections for each cluster. To quantify this effect, we present both the mean and variance of the individual detections in each cluster for the five case studies in Figures~\ref{fig:means} and \ref{fig:vars}, respectively.

\begin{figure*}[!ht]
\centering
\begin{subfigure}{0.3145\linewidth}
\centering
\includegraphics[width=\textwidth]{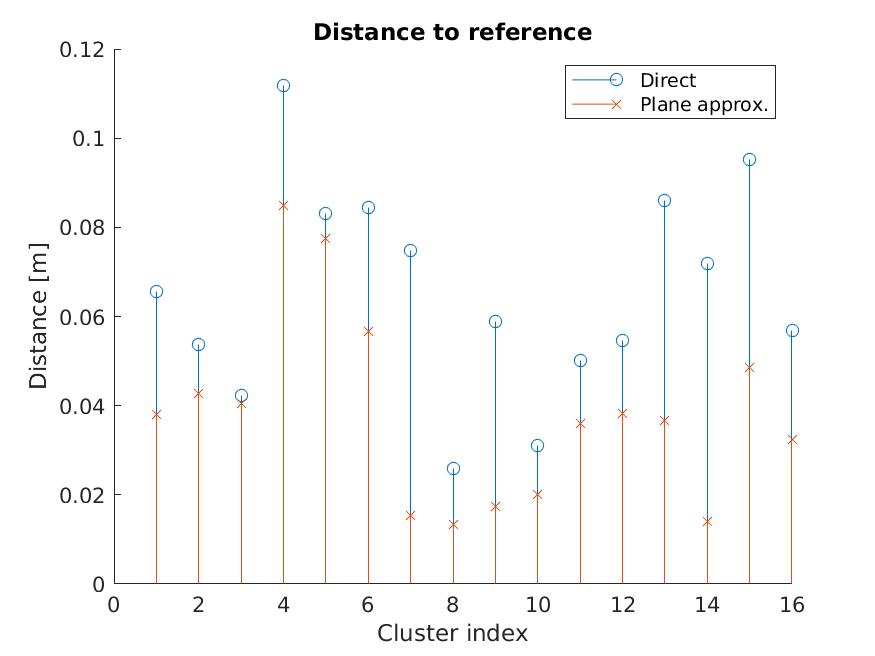}
\caption{}
\end{subfigure}
\begin{subfigure}{0.3145\linewidth}
\centering
\includegraphics[width=\textwidth]{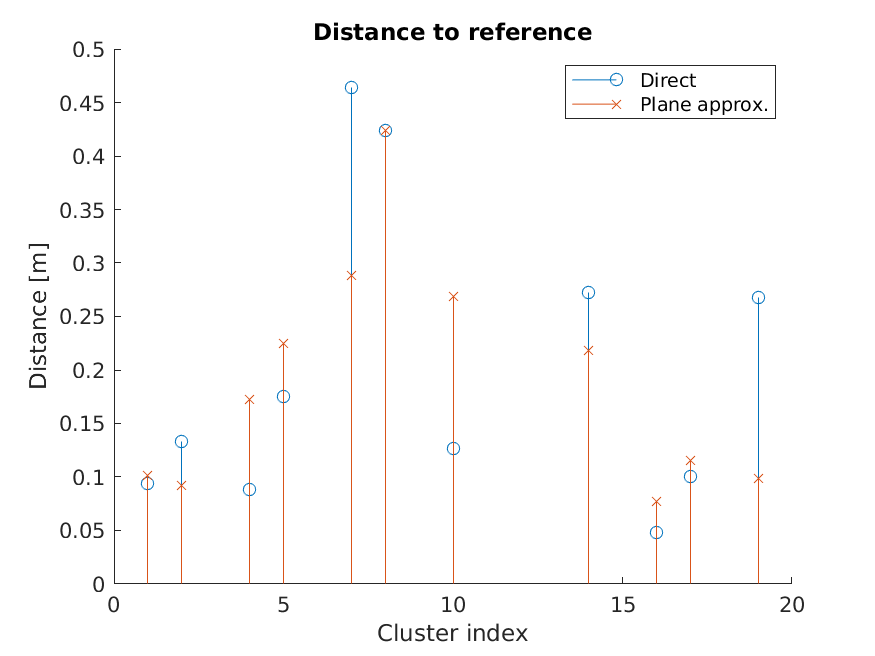}
\caption{}
\end{subfigure}
\begin{subfigure}{0.3145\linewidth}
\centering
\includegraphics[width=\textwidth]{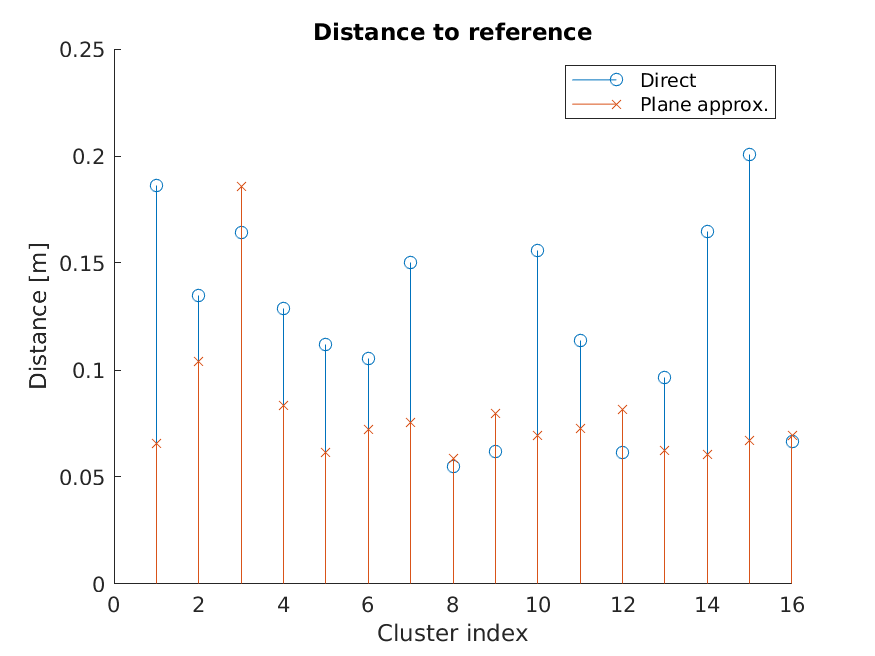}
\caption{}
\end{subfigure}
\begin{subfigure}{0.3145\linewidth}
\centering
\includegraphics[width=\textwidth]{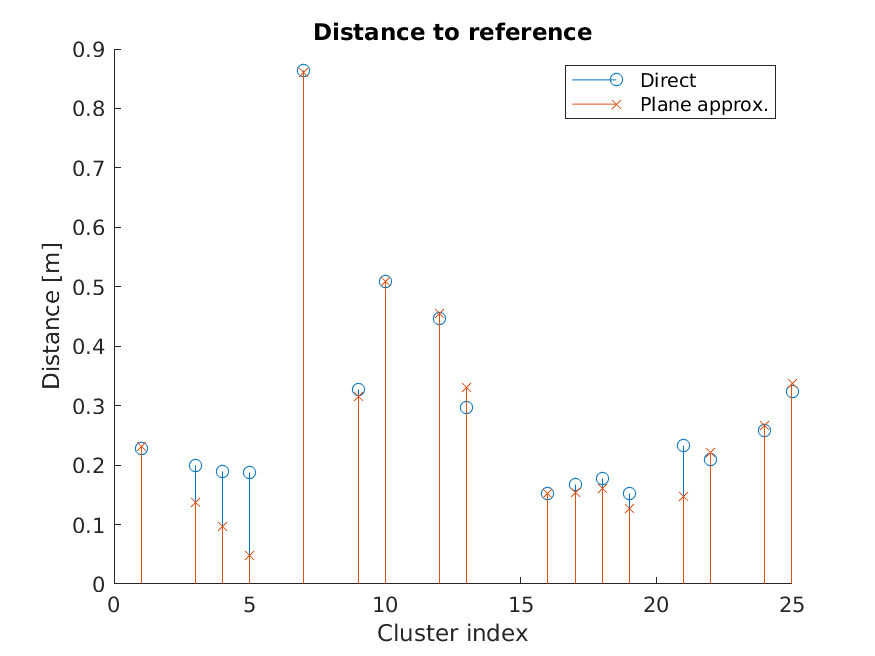}
\caption{}
\end{subfigure}
\begin{subfigure}{0.3145\linewidth}
\centering
\includegraphics[width=\textwidth]{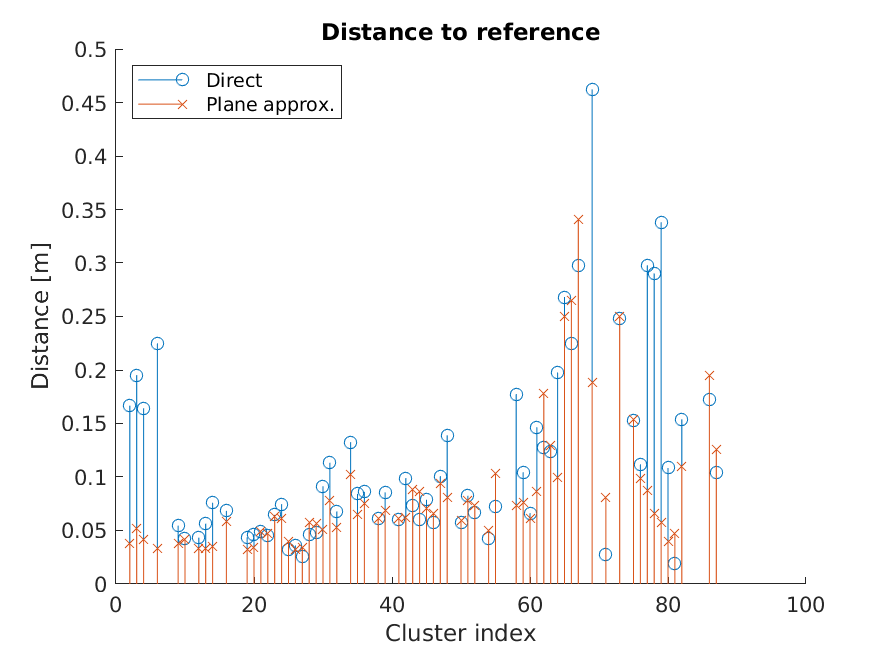}
\caption{}
\end{subfigure}
\begin{subfigure}{0.3145\linewidth}
\centering
\includegraphics[width=\textwidth]{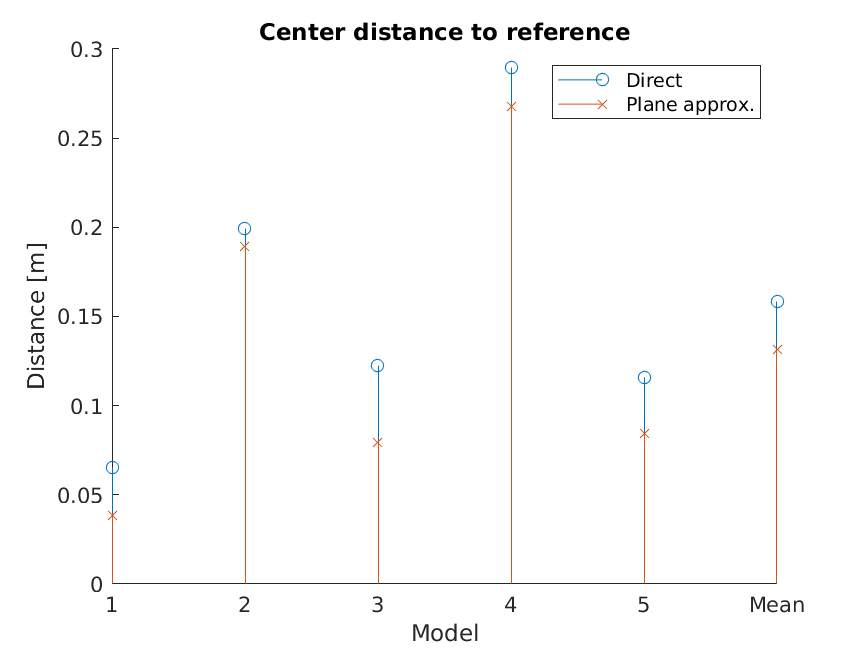}
\caption{}
\end{subfigure}
\caption{Cluster center distances to reference with and without plane estimation and filtering. (a-e) Models 1-5; (f) global statistics.}
\label{fig:rdis}
\end{figure*}

We can see that the average values are almost always lower with plane estimation and filtering, with a total average for all the detections decreasing from 10.93 cm to 7.98 cm for the mean and 638 cm$^2$ to 325 cm$^2$ for the variance.
More results are presented in Figure~\ref{fig:rdis}, showing the distance between cluster centers and reference positions. In this case, the average distance is lower for the five case studies when plane estimation is used, from 15.86 cm to 13.17 cm on average.

\begin{figure*}[!ht]
\centering
\begin{subfigure}{0.315\textwidth}
\centering
\includegraphics[width=\textwidth]{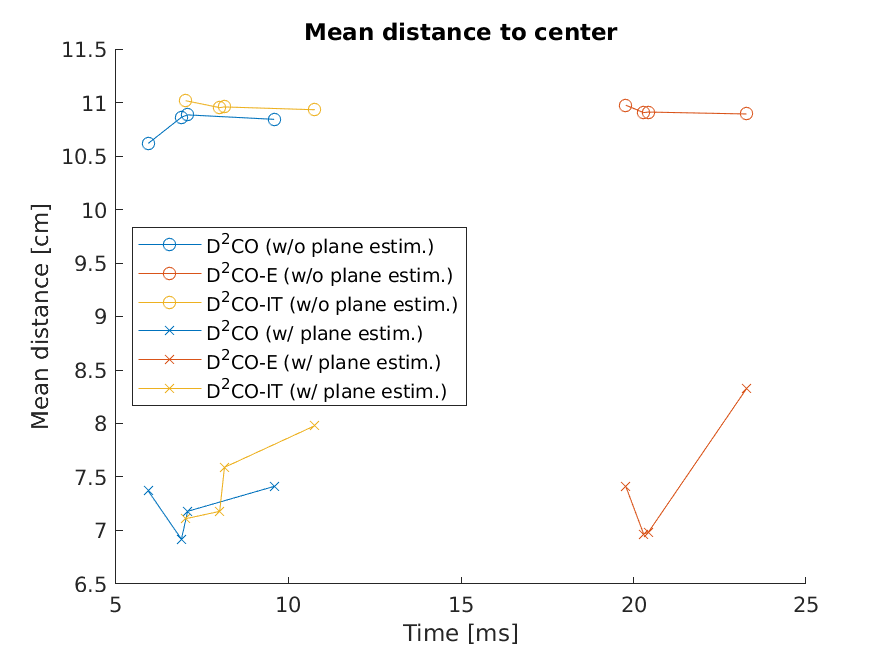}
\caption{}
\end{subfigure}
\begin{subfigure}{0.315\textwidth}
\centering
\includegraphics[width=\textwidth]{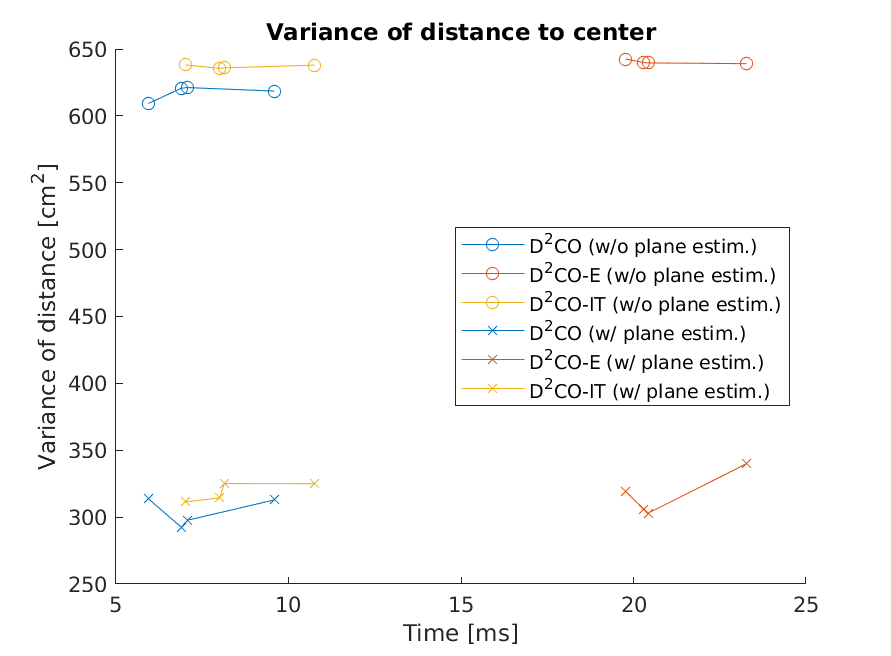}
\caption{}
\end{subfigure}
\begin{subfigure}{0.315\textwidth}
\centering
\includegraphics[width=\textwidth]{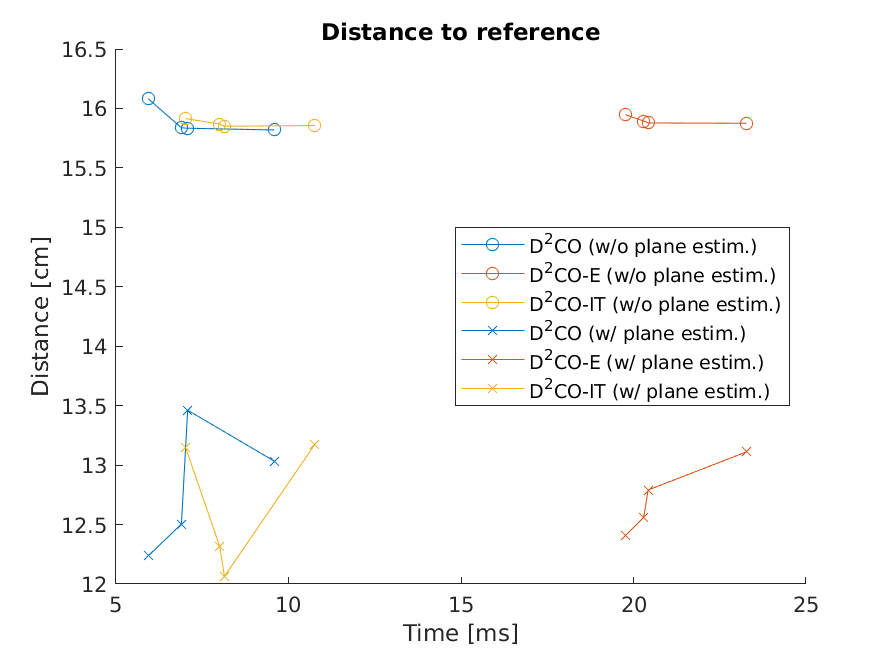}
\caption{}
\end{subfigure}
\caption{Localization statistics with (w/) and without (w/o) plane estimation and filtering for different optimization methods with step values of 25\%, 50\%, 75\% and 100\% of the largest edge in the model.}
\label{fig:times_loc}
\end{figure*}

Finally, Figure~\ref{fig:times_loc} presents localization statistics for different optimization methods with and without plane estimation, with some numerical values in Table~\ref{tab:times_loc}. As previously mentioned, the use of the plane estimation step improves the global results; however, in this case, the localization results are very similar for all the optimization methods.

\begin{table*}[!ht]
\centering
\small
\caption{Average localization statistics over step values of 25\%, 50\%, 75\% and 100\% for different optimization methods with (w/) and without (w/o) plane estimation and filtering.}
\begin{tabular}{lccc}
\hline
 & \textbf{D$^2$CO} & \textbf{D$^2$CO-E} & \textbf{D$^2$CO-IT} \\
\hline
Time [ms] & 8.2387 & 21.8801 & 9.4107 \\
Mean dist. to center (w/o p. e.) [cm] & 10.7809 & 10.9150 & 10.9515 \\
Mean dist. to center (w/  p. e.) [cm] & 7.3147 & 7.4200 & 7.2568 \\
Var. of dist. to center (w/o p. e.) [cm$^2$] & 619.2060 & 639.7888 & 636.4619 \\
Var. of dist. to center (w/  p. e.) [cm$^2$] & 309.9921 & 316.9307 & 312.4581 \\
Distance to reference (w/o p. e.) [cm] & 15.8886 & 15.8964 & 15.8613 \\
Distance to reference (w/  p. e.) [cm] & 12.6824 & 12.6800 & 12.5438 \\
\hline
\end{tabular}
\label{tab:times_loc}
\end{table*}

\section{Conclusions}
\label{sec:conclusions}

The system presented in this work provides an automatic solution for detecting, identifying and localizing lighting elements with improved accuracy with respect to previous works from the authors~\cite{Troncoso1, Troncoso2} thanks to a new optimization method and a better use of the available BIM information of a building. This detection system was applied to five case studies with a total of 166 lamps of different models and more than 30,000 images in the dataset.

The experimental results show an improvement in each of the key parts of the process. First, there are 79.4 more detections on average without plane estimation and 93.1 with plane estimation. Second, the identification is 100\% correct for the 65 detected clusters, and the identification for the individual detections is improved from 1.37\% to 0.43\% of error. Finally, the dispersion of the positions is reduced, with distances from detections to cluster centers decreasing from 10.78 cm and 619.21 cm$^2$ to 7.26 cm and 312.46 cm$^2$ for the mean and variance, respectively. Moreover, the average distance between cluster centers and reference positions is also reduced from 15.89 cm to 12.54 cm. These results evidence the validity of the system and the new enhancements, with better results in terms of detection, identification and localization.

In this work, we leverage valuable information from the BIM to improve the detection system, but more can be done in this regard. We are currently evaluating the use of BIM information also on the first steps of the process to further increase the accuracy and reliability of the system.

\section{Acknowledgements}
\label{sec:ack}

This work is funded by the European Regional Development Fund (ERDF)
and the Galician Regional Government under agreement for funding the
Atlantic Research Center for Information and Communications Technologies
(AtlantTIC), the Spanish Ministry of Economy and Competitiveness under
the National Science Program (TEC2014-54335-C4-3-R and
TEC2017-84197-C4-2-R). This investigation article was also partially supported by the INMENA project through the Xunta de Galicia CONECTA PEME 2018 (IN852A 2018/59). Moreover, the authors want to give thanks to the Xunta de Galicia (Grant ED481A).


%
\bibliographystyle{ieeetr} 
\bibliography{biblio}

\end{document}